\pdfoutput=1
\documentclass[]{style}

\titlecardtrue

\title{Thinking with Images without Coordinates}

\toplogo[trim=12.23bp 12.38bp 18.98bp 12.23bp, clip]{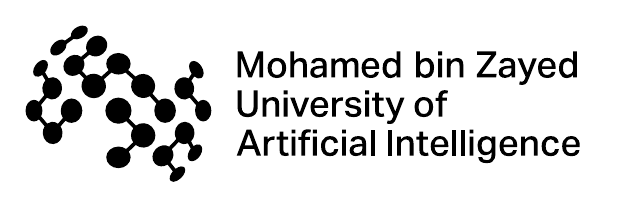}

\author[1${*}$]{Meng Cao}
\author[1${*}$]{Haoze Zhao}
\author[2]{Yi Wang}
\author[1]{Ian Reid}
\author[1,3]{Xiaodan Liang}

\affiliation[1]{Mohamed bin Zayed University of Artificial Intelligence}
\affiliation[2]{Shanghai AI Lab}
\affiliation[3]{Sun Yat-sen University}

\metadata[Project Page]{\url{https://mengcaopku.github.io/TiCO/}}
\metadata[GitHub]{\url{https://github.com/mengcaopku/TiCO}}

\usepackage{amsmath}
\usepackage{amssymb}
\usepackage{array}
\usepackage{longtable}
\usepackage{float}
\usepackage{enumitem}
\usepackage{xspace}
\usepackage{wrapfig}
\usepackage{makecell}
\usepackage{pifont}
\usepackage{arydshln}

\IfFileExists{algorithm.sty}{\usepackage{algorithm}}{}
\IfFileExists{algpseudocode.sty}{\usepackage{algpseudocode}}{}
\newcommand{\ie}{\textit{i.e.}\xspace}
\newcommand{\eg}{\textit{e.g.}\xspace}

\newcommand{\cf}{\textit{cf.}\xspace}
\newcommand{\vs}{vs.\xspace}

\newcommand{\answer}[1]{\textcolor{purple}{\texttt{<answer>}} #1 \textcolor{purple}{\texttt{</answer>}}}

\abstract{
Recent advances in multi-modal large language models have shifted toward the ``thinking with images" paradigm to improve reasoning faithfulness, yet existing methods rely heavily on explicit \emph{coordinate-based} grounding. We identify two fundamental limitations in this approach, \ie, a semantic gap where sparse regressions miss high-dimensional latent structures, and a resolution gap due to the granularity mismatch between patch-level tokens and pixel-level coordinates. To address these, we propose \textbf{TiCO} (\textbf{T}hinking with \textbf{i}mages without \textbf{CO}ordinates). It reformulates visual reasoning as a semantic evidence discovery process within the latent space, and evidence is injected as token-level regions without pixel coordinates. Specifically, TiCO utilizes a dedicated probe token to estimate similarity distributions and performs importance sampling over spatially connected components to generate diverse, semantically dominant evidence rollouts. Extensive experiments across multiple high-resolution and general visual reasoning benchmarks demonstrate that our TiCO consistently outperforms state-of-the-art coordinate-based methods with significantly improved inference efficiency.
}

\begin{document}

\maketitle
\vspace{1mm}
\begin{center}
	   %\vspace{-0.2cm}
		\includegraphics[width=\linewidth]{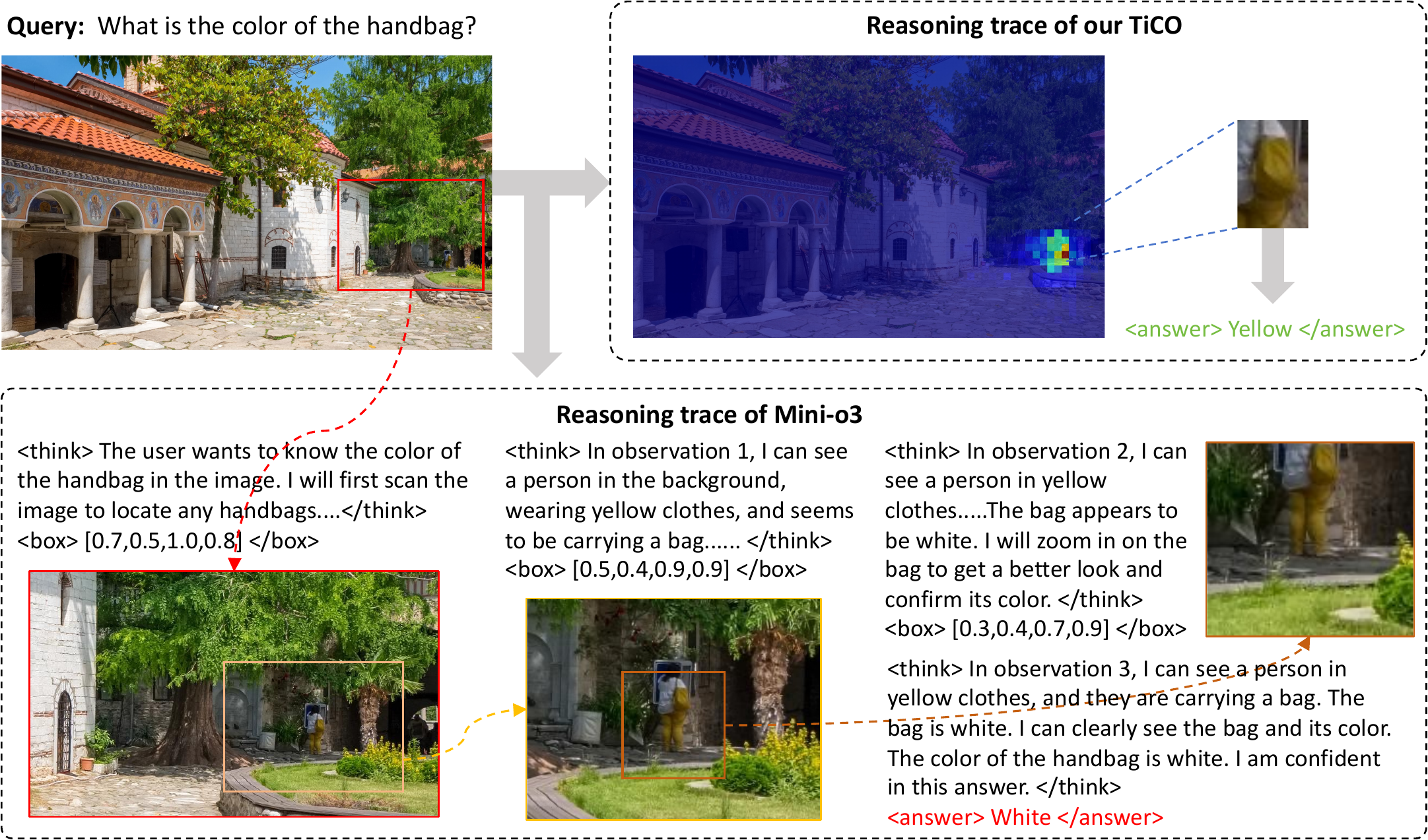} 
		\captionsetup{type=figure}
		%\vspace{-0.1cm}
     %\caption{\textbf{Comparison between coordinate-based method and our TiCO.} Traditional models like Mini-o3 (bottom) rely on explicit coordinate regression (\texttt{<box>}), which is highly sensitive to numerical precision and prone to error propagation. In contrast, our proposed TiCO (top) shifts the focus from explicit coordinate regression to latent semantic region indexing. By decoupling reasoning from numerical coordinates, TiCO directly attends to informative regions within the latent space, maintaining superior contextual consistency and successfully identifying the small \texttt{yellow handbag} where coordinate-dependent methods fail.}
     \caption{\textbf{Comparison between coordinate-based methods and TiCO.} Traditional model Mini-o3 \cite{lai2025mini} relies on explicit coordinate regression (\texttt{<box>}), which is sensitive to localization errors and prone to error propagation. In contrast, our TiCO replaces coordinate prediction with latent semantic region indexing, directly attending to informative regions in the token space.}
     % thereby maintaining stronger contextual consistency and correctly identifying the small \texttt{yellow handbag} where coordinate-based methods fail.
	%\vspace{-0.2cm}
	\label{fig:teaser}
\end{center}

% Figure 1: Comparison of reasoning paradigms between the coordinate-based approach (e.g., Mini-o3) and our proposed TiCO (Thinking with Images without Coordinates). Given the query "What is the color of the helmet?", traditional models like Mini-o3 (bottom) rely on an iterative "zoom-and-crop" loop, requiring the explicit generation of numerical bounding box coordinates (<box>). This process is prone to error accumulation; as shown, a slight misalignment in the final crop leads to an incorrect answer ("Red"). In contrast, TiCO (top) performs visual reasoning directly within the latent space. By leveraging an inherent "thinking" process that bypasses the need for explicit coordinate prediction, TiCO maintains a global-to-local context more effectively, successfully identifying the small green helmet in the distant background. Our method demonstrates superior efficiency and accuracy by eliminating the overhead and sensitivity associated with precise coordinate regression in complex multi-step visual reasoning tasks.
\section{Introduction} \label{sec:intro}

\begin{figure*}[t]
	\centering
    \includegraphics[width=\textwidth]{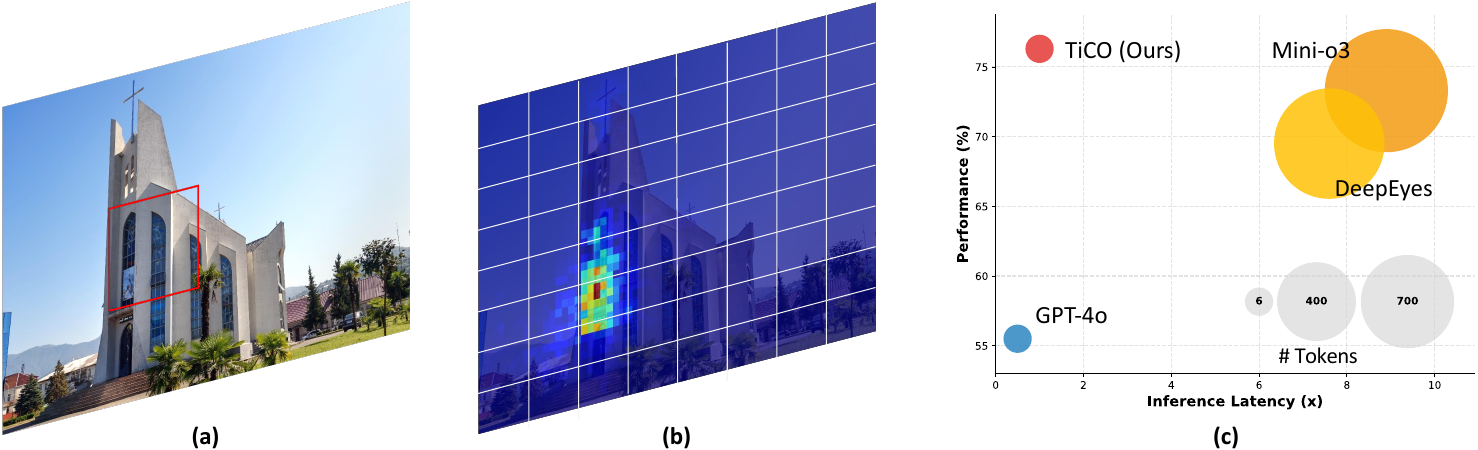}
    % \vspace{-1mm}
	\caption{\textbf{Conceptual comparison and performance trade-offs.} \textbf{(a) Coordinate-based methods} \cite{lai2025mini,zheng2025deepeyes} localize evidence via explicit bounding-box prediction, reducing reasoning to geometric regression in image space; \textbf{(b) Coordinate-free methods} instead operate directly in visual token space, inducing a similarity distribution over patches and extracting semantically coherent evidence without explicit coordinates; \textbf{(c) Performance–efficiency comparison} on HR-Bench 8K \cite{wang2025divide}. Our TiCO achieves superior accuracy with substantially lower inference latency and fewer tokens compared to coordinate-based approaches such as Mini-o3 \cite{lai2025mini} and DeepEyes \cite{zheng2025deepeyes}.}
	\label{fig:ConceptCompare}
	%\vspace{-2mm}
\end{figure*}

Multi-modal Large Language Models (MLLMs) \cite{gpt4o,anthropic2024claude,bai2025qwen2} have emerged as powerful general-purpose agents to interpret and execute diverse real-world tasks. Beyond basic perception, recent advances increasingly position MLLMs as reasoning systems \cite{wang2025multimodal,huang2023towards,xu2025towards} that must interpret visual scenes, integrate contextual knowledge, and generate structured responses. Early approaches predominantly adopt \emph{text-centric} reasoning pipelines \cite{yao2023tree,wei2022chain,zhang2023multimodal,xu2024llava,thawakar2025llamav,yao2024mulberry}, where visual inputs are compressed into latent embeddings and subsequent reasoning unfolds largely within the language domain. This design weakens the alignment between reasoning processes and grounded visual evidence, resulting in reduced reliability and interpretability \cite{gupta2023bias,xu2024pride,zhou2024unibias}. To mitigate this issue, recent efforts shift toward interleaved multimodal reasoning, \ie, \emph{thinking with images} \cite{shao2024visual,su2025thinking,lai2025mini,zheng2025deepeyes}, which explicitly anchors intermediate reasoning to semantically grounded visual regions (\cf Fig. \ref{fig:teaser}), thereby improving faithfulness and robustness.

Despite its advantages, most existing thinking-with-images methods \cite{lai2025mini,zheng2025deepeyes,jiang2025rex,sarch2025grounded,wang2025traceable,wang2025vgr,bai2025multi,cao2025ground} rely on explicit coordinate prediction for grounding. As shown in Fig. \ref{fig:ConceptCompare}(a), given an image–query pair, the model typically predicts spatial coordinates and then conditions subsequent reasoning on the cropped visual region. While intuitive, this coordinate-based design presents several limitations. \emph{1) Semantic Gap}: Grounding is reduced to low-dimensional numerical regression (\eg, bounding boxes), which fails to capture the rich semantic structures encoded in the high-dimensional latent space. Moreover, MLLMs are fundamentally trained under a next-token-prediction paradigm, where numerical tokens are modeled as \emph{discrete} symbols rather than \emph{continuous} geometric quantities. As a result, precise coordinate generation is not naturally aligned with their training objective \cite{chen2021pix2seq,li2025discrete,peng2023kosmos}, and geometric correctness does not necessarily imply faithful semantic grounding. \emph{2) Resolution Gap}: A granularity mismatch exists between visual representations and coordinate outputs. Vision encoders such as ViT \cite{dosovitskiy2020image} operate on discrete patch-level tokens with coarse spatial granularity, whereas coordinate prediction requires continuous pixel-level precision. This discrepancy forces MLLMs to generate fine-grained numerical locations from coarse visual tokens, further increasing learning difficulty and instability.

%As shown in Fig. \ref{fig:ConceptCompare}(a), given an image–query pair, the model typically generates spatial coordinates and subsequently conditions reasoning on the cropped visual evidence. While intuitive, this design introduces several limitations: \emph{1) Semantic Gap}: The conventional coordinate-based paradigm constrains visual grounding to low-dimensional numerical regressions (\eg, bounding boxes). Such sparse representations do not explicitly encode the underlying semantic reasoning structure within the high-dimensional latent space. As a result, geometric correctness does not necessarily guarantee faithful alignment with the semantically relevant visual evidence required for reasoning; \emph{2) Resolution Gap}: There exists an inherent granularity inconsistency between visual representations and coordinate-based outputs. Modern vision encoders, \eg, ViT \cite{dosovitskiy2020image}, operate on discrete \emph{patch-level} tokens that provide coarse spatial representations. In contrast, coordinate prediction requires continuous, high-resolution \emph{pixel-level} outputs. This discrepancy forces MLLMs to regress continuous geometric locations from visual features, introducing unnecessary learning difficulty.

To this end, we propose \textbf{T}hinking with \textbf{i}mages without \textbf{CO}ordinates (\textbf{TiCO}), a coordinate-free paradigm that grounds visual reasoning directly in semantic token interactions, eliminating the need for explicit coordinate-based localization. We argue that coordinate prediction is an ill-posed interface. Instead of predicting explicit spatial coordinates, we prompt MLLMs to output a special token \texttt{<GROUND>}, which acts as a semantic probe to estimate a similarity distribution over visual tokens. Based on the similarity map, we first aggregate token-level attentive scores within spatially connected regions to compute a confidence score for each connected component. We then perform importance sampling over these connected components according to their aggregated scores, generating a set of \emph{evidence rollouts} that prioritize semantically dominant regions while preserving distributional diversity. Conditioned on the sampled evidence regions, the model subsequently performs \emph{answer rollouts}. In this way, reasoning trajectories are grounded not in explicit geometric coordinates, but in semantically sampled evidence regions derived directly from token-level similarity distributions. Furthermore, by eliminating redundant textual reasoning steps, our approach offers significant efficiency gains over prior methods. As shown in Fig. \ref{fig:ConceptCompare}(c), on HR-Bench 4K \cite{wang2025divide}, our TiCO surpasses DeepEyes \cite{zheng2025deepeyes} by 4.8\% in overall accuracy while using only 1/64 of the inference tokens and 1/9 of the inference time. %\textcolor{red}{Need to Mention the Optimization}

Our contributions are summarized as follows:
\begin{itemize}[topsep=0pt, partopsep=0pt, leftmargin=13pt, parsep=0pt, itemsep=3pt]
    \item We introduce TiCO, a novel framework for thinking with images without coordinates, which reformulates visual reasoning as semantic evidence discovery in visual token space instead of numeric bounding box regression. 
    
    \item TiCO formulates evidence discovery as a similarity-induced region sampling process, where a semantic probe token interacts with visual patches to generate distribution-guided connected components.

    \item Extensive experiments demonstrate that our approach achieves superior performance on both high-resolution and general visual reasoning benchmarks while substantially reducing inference overhead. %\textcolor{red}{On HR-Bench, it surpasses Mini-o3 by +5.3\% in average accuracy with only one-fifth of the inference cost.}
\end{itemize}

\section{Related Work} \label{sec:related}

\noindent \textbf{Reasoning-oriented MLLMs.} The groundbreaking success of advanced reasoning LLMs \cite{wang2025multimodal,huang2023towards,xu2025towards}, exemplified by systems like OpenAI-o1 \cite{jaech2024openai} and DeepSeek-R1 \cite{guo2025deepseek}, has motivated efforts to extend human-level cognitive competencies to multi-modal settings. Pioneering works \cite{chen2023see, yao2023tree, besta2024graph, wei2022chain} utilize deliberate prompting strategies to facilitate the emergence of explicit reasoning rationales during inference. Subsequent efforts enhance reasoning through structured SFT data regimes, such as defining explicit multi-stage rationales in LLaVA-CoT \cite{xu2024llava} or integrating Monte Carlo Tree Search with self-reflective feedback for iterative self-correction in Mulberry \cite{yao2024mulberry}. Recent R1-inspired methodologies \cite{yang2025r1,peng2025lmm,tan2025reason,liu2025visual,meng2025mm,yu2025perception,huang2025vision} have demonstrated that long-chain reasoning capabilities in MLLMs can be effectively elicited solely through the application of answer-based rewards. Vision-R1 \cite{huang2025vision} introduces a progressive thinking suppression strategy to mitigate overthinking artifacts, while Visual-RFT \cite{liu2025visual} demonstrates that the robust reasoning can be generalized to specialized vision tasks such as fine-grained classification and object detection. Despite these advancements, most LMMs still rely on a text-centric reasoning paradigm that lacks the interleaved visual-textual grounding inherent in human cognition, necessitating a shift toward more visually-anchored reasoning frameworks.

\noindent \textbf{Thinking with Images.} A recent milestone, OpenAI-o3 \cite{o3}, advanced multi-modal reasoning by enabling dynamic image manipulation, such as cropping and zooming, to emulate human-like thinking-with-images. To empower open-source models with such interleaved reasoning capabilities, recent efforts \cite{shao2024visual,su2025thinking,lai2025mini,zheng2025deepeyes,wang2025traceable,wang2025vgr,zhu2025active} have leveraged either vanilla RL or RL with cold-start SFT to replicate the successes of proprietary models. Typically, these methodologies rely on numerical bounding box generation to facilitate focusing on region-of-interest. However, coordinate-based regression often introduces significant bottlenecks: numerical tokens suffer from a granularity mismatch with high-resolution visual features, and explicit coordinates are highly sensitive to quantization errors, which can lead to error accumulation during multi-step reasoning. While GUI-Actor \cite{wu2025gui} similarly adopts similarity-based selection to bypass coordinate regression, it remains distinct from our work in several aspects: \emph{1) Domain limitation}: It is specialized for structured GUI elements rather than complex, open-ended natural scenes; 2) \emph{Optimization gap}: Its reliance on an SFT-only paradigm and requires external verifiers for selected region verification, which increases computational overhead; and 3) \emph{Task focus}: It prioritizes atomic grounding for actions over the multi-step visual reasoning required for sophisticated problem-solving. 

% \textcolor{red}{=======}
% % High-resolution benchmarks
% \textcolor{red}{=======}
\section{Method} \label{sec:method}

\begin{figure*}[t]
	\centering
    \includegraphics[width=\textwidth]{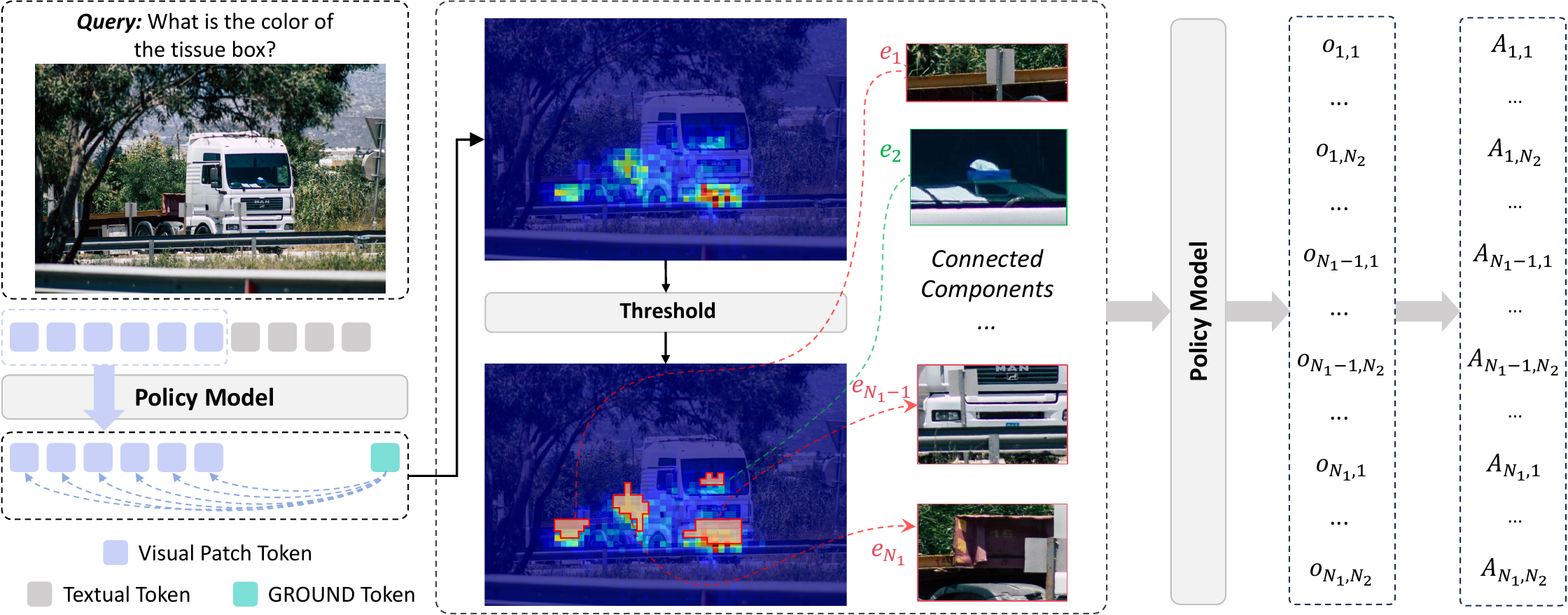}
    % \vspace{-1mm}
	\caption{\textbf{Overview of TiCO.} Given an image–text query pair, the policy model first produces a semantic probe token (\texttt{<GROUND>}) whose embedding interacts with visual patch tokens to generate a similarity map. After thresholding the similarity distribution, spatially connected components are extracted as candidate evidence regions. These components are treated as evidence rollouts and are subsequently fed back into the policy model to generate answer rollouts.}
	\label{fig:pipeline}
	%\vspace{-2mm}
\end{figure*}

The schematic illustration of our TiCO is illustrated in Fig. \ref{fig:pipeline}. We first introduce a probe-based anchoring mechanism (Sec. \ref{sec:3.1}) to identify task-relevant regions in the latent semantic space, followed by region-grounded answering (Sec. \ref{sec:3.2}) to integrate this visual evidence into an interleaved reasoning trajectory. Finally, we detail our optimization strategy (Sec. \ref{sec:3.3}).

\subsection{Probe-based Anchoring} \label{sec:3.1}

Existing thinking-with-images approaches rely on explicit spatial coordinates to localize regions for visual reasoning, introducing rigid geometric constraints and often requiring additional supervision. In contrast, we propose a coordinate-free framework in which a semantic probe token directly interacts with visual representations. Visual tokens with higher semantic similarity are preferentially selected to form evidence rollouts, which are then incorporated into the reasoning process to produce the final answer rollouts.

Given an image $I$ and a textual query $Q$, a MLLM encodes the image into a sequence of visual tokens $\mathbf{V} = \{v_1, \dots, v_N\}$, where each token $v_i \in \mathbb{R}^d$ corresponds to a spatial patch in the image and $d$ denotes the embedding dimension. The textual query is tokenized into a sequence $\mathbf{T} = \{t_1, \dots, t_M\}$. To enable coordinate-free grounding, we prompt MLLM to generate a semantic anchor \texttt{<GROUND>} instead of the specific coordinates. Then, the final-layer hidden state of \texttt{<GROUND>}, \ie, $h_{\texttt{<GROUND>}} \in \mathbb{R}^d$ serves as a semantic anchor to filter the salient visual patches. Specifically, we compute the similarity between the embedding of probe token $h_{\texttt{<GROUND>}}$ and each visual token $v_i$ using scaled dot-product similarity and yielding a similarity map over visual tokens. 
\begin{equation}
    \alpha_i = \frac{\exp(s_i)}{\sum_{j=1}^{N} \exp(s_j)},
\quad s_i = \frac{h_{\texttt{<GROUND>}}^\top v_i}{\sqrt{d}},
\end{equation}
where $\alpha_i$ reflects the likelihood of the token $v_i$ contributing to the reasoning process conditioned on the probe token. 

\subsection{Region-Grounded Answering} \label{sec:3.2}

\noindent \textbf{Evidence Region Rollout.} Based on the attention score distribution $\{\alpha_i\}_{i=1}^{N}$, we apply a threshold $\tau$ to derive a binary mask over visual tokens:
\begin{equation}
\mathcal{M} = \{ v_i \mid \alpha_i > \tau \}.
\label{eq:2}
\end{equation}
Then, we compute connected components on this binary map based on spatial adjacency, resulting in a set of candidate regions $\{\mathcal{C}_1, \dots, \mathcal{C}_K\}$.

To determine the final evidence rollouts, we evaluate each connected component according to the aggregated attention mass of the visual tokens it contains. For each component $\mathcal{C}_k$, we compute its attention mass and normalize it into a probability distribution over connected components in a unified form:
\begin{equation}
P(\mathcal{C}_k) = 
\frac{\exp(M_k)}
{\sum_{j=1}^{K} \exp(M_j)},
\qquad
M_k = \sum_{v_i \in \mathcal{C}_k} \alpha_i,
\end{equation}
\noindent where $M_k$ denotes the aggregated attention weight for each connected component, $P(\mathcal{C}_k)$ represents the normalized region-level probability distribution.

Subsequently, we employ \emph{importance sampling} to select $N_1$ evidence regions from the underlying distribution, ensuring that highly relevant semantic anchors are prioritized for the reasoning process. Unlike deterministic \emph{top-scoring} selection, which greedily focuses on the highest-scoring components and may over-commit to potentially noisy peaks, importance sampling maintains stochastic exploration over semantically plausible regions. This is particularly beneficial under reinforcement learning, where diverse evidence rollouts provide richer trajectory signals and mitigate premature convergence caused by suboptimal but locally dominant regions.
\begin{equation}
\mathcal{R}_{i} \sim P(\mathcal{C}_k), \quad n = 1, \dots, N_1,
\end{equation}
\noindent where $\mathcal{R}{i}$ denotes the cropped region corresponding to the minimal bounding rectangle enclosing the connected component.

\noindent \textbf{Answer Rollout.} Leveraging the sampled evidence region rollouts $\{\mathcal{R}_{n}\}_{n=1}^{N_1}$, we condition the policy model $\pi_\theta$ on the visual evidences to synthesize final responses. Formally, for the $n$-th evidence region rollout, we generate $N_2$ answer rollouts according to the conditional policy:
\begin{equation}
\mathbf{o}_{i,j} 
\sim 
\pi_\theta(\cdot \mid Q, \mathbf{V}, \mathcal{R}_{i}), 
\quad 
j \in [1, N_2], 
\; i \in [1, N_1],
\label{eq:5}
\end{equation}
where $\mathbf{o}_{i,j}$ denotes the $j$-th answer rollout conditioned on the $i$-th evidence region rollout. $N_2$ denotes the number of answer rollouts generated per evidence region, resulting in a total of $N_1 \cdot N_2$ reasoning trajectories.

\subsection{Optimization} \label{sec:3.3}

Our optimization follows a two-stage paradigm. For SFT, we train the model to generate semantic probe tokens and employ dense patch-wise supervision to ensure these tokens precisely attend to task-relevant regions. Then, we utilize GRPO \cite{shao2024deepseekmath} in the RL stage to further incentivize the visual reasoning capability.

\noindent \textbf{Supervised Fine-tuning.} Instead of relying on sparse bounding-box-level supervisions in previous coordinate-based methods, we introduce a dense supervision on the predicted similarity map by treating all image patches that are partially or fully covered by the ground-truth bounding box as positive examples and all others as negatives.
\begin{equation}
\mathcal{L}_{\text{sft}}^{\text{ground}}
= -\sum_{i=1}^{N} \hat{\alpha}_i \log \alpha_i,
\label{eq:6}
\end{equation}
\noindent where $\hat{\alpha}_i$ denotes the binary label for the corresponding visual patch. The overall training objective is composed of both $\mathcal{L}_{\text{sft}}^{\text{ground}}$ and the standard next-token prediction loss $\mathcal{L}_{\text{sft}}^{\text{NTP}}$:
\begin{equation}
\mathcal{L}_{\text{sft}} = \mathcal{L}_{\text{sft}}^{\text{ground}} + \mathcal{L}_{\text{sft}}^{\text{NTP}}.
\end{equation}

\noindent \textbf{Reinforcement Learning.} TiCO is trained using GRPO \cite{shao2024deepseekmath}, enabling joint RL over evidence grounding and answer generation. We adopt a reward formulation including an accuracy reward and a formatting reward. For format reward, we constrain the \texttt{<GROUND>} token bracketed within \textcolor{blue}{\texttt{<GROUND$\_$START>}} and \textcolor{blue}{\texttt{<GROUND$\_$End>}} and the answer within \answer{and}. Following Video-R1 \cite{feng2025video}, the answer reward is defined as the lexical alignment using the average of ROUGE-1, ROUGE-2, and ROUGE-L scores against ground-truth answers. The total reward for the response $\boldsymbol{o}_{i,j}$ is denoted as ${r}_{i,j}$.
\begin{equation}
\begin{gathered}
\mathcal{J}_{G R P O}(\theta)=\mathbb{E}\left[I, Q,\left\{\boldsymbol{o}_{i}\right\}_{i=1,j=1}^{N_1,N_2} \sim \pi_{\theta_{\text {old}}}(\boldsymbol{o} \mid I, Q)\right] \frac{1}{N_1N_2} \sum_{i=1}^{N_1}\sum_{j=1}^{N_2} \\
\min \left(\frac{\pi_\theta(\boldsymbol{o}_{i,j} \mid I, Q)}{\pi_{\theta_{\text{old}}}(\boldsymbol{o}_{i,j} \mid I,Q)} A_{i,j}, \operatorname{clip}\left(\frac{\pi_\theta\left(\boldsymbol{o}_{i,j} \mid I, Q\right)}{\pi_{\theta_{old}}\left(\boldsymbol{o}_{i,j} \mid I,Q\right)}, 1-\varepsilon, 1+\varepsilon\right) A_{i,j}\right),
\end{gathered}
\label{eq:3}
\end{equation}
\noindent where $\pi_\theta$ denotes the current policy model. $\boldsymbol{o}_{i,j}$ denote the $j$-th rollout answers based on the $i$-th evidence region (\cf Eq \eqref{eq:5}). $\varepsilon$ are hyper-parameters. $A_{i,j}$ are the advantages computed as follows.
% using the group reward $\{r^{\text{answer}}_{i,j} \mid 1 \leq i \leq G_1, \ 1 \leq j \leq G_2\}$
%-----------------------------------------------------------------------
%-----------------------------------------------------------------------
\begin{equation}
A_{i,j}=\frac{r_{i,j}-\operatorname{mean}(\{r_{i,j}\}_{i=1,j=1}^{N_1,N_2})}{\operatorname{std}(\{r_{i,j}\}_{i=1,j=1}^{N_1,N_2})}. \label{eq:4}
\end{equation}
%-----------------------------------------------------------------------

\noindent \textbf{Inference.} During inference, the MLLM is prompted to generate the \texttt{<GROUND>} probe token to attend to relevant visual patches when necessary. Based on the resulting similarity distribution, we form connected components and sample evidence regions without performing any explicit coordinate regression. The model then conditions on the selected regions to generate the final answer. The entire process remains coordinate-free and operates directly in the visual token space.
\section{Experiments} \label{sec:experiments}

\subsection{Setups}

\noindent \textbf{Training Datasets.} For SFT, we adapt the cold-start SFT dataset introduced in Mini-o3 \cite{lai2025mini} to accommodate our dense similarity-map supervision. Since the Mini-o3 SFT data contain multi-step refined bounding box predictions, we directly take the final-round bounding box as the ground-truth region. All visual patches that are partially or fully covered by this ground-truth box are labeled as positive. Please refer to the supplementary material for more details. For RL, we directly adopt the RL dataset constructed by Mini-o3 for training.

\noindent \textbf{Implementation Details.} Our implementation is based on Qwen-2.5-VL-7B-Instruct \cite{bai2025qwen2}. The training process consists of a SFT phase ($640$ steps, batch size $8$, learning rate $5 \times 10^{-6}$) followed by a RL phase ($4000$ steps, batch size $8$, learning rate $1 \times 10^{-6}$). For the multi-stage inference, we set the rollout counts to $N_1 = 4$ for grounding and $N_2 = 2$ for answering, utilizing an evidence selection threshold of $\tau = 0.15$. During generation, a default sampling temperature of $1$ and a maximum response length of $512$ tokens are applied. The entire framework was trained on $8 \times$ H100 GPUs, totaling approximately $12$ hours of computation.
% \textcolor{red}{TODO} We conducted experiments based on Qwen-2.5-VL-7B-Instruct \cite{bai2025qwen2}. The SFT training was performed for 640 steps with a batch size of 8 and a learning rate of {$5\times e^{-6}$}. The RL training lasts for 4000 steps with a batch size of 8 and a learning rate of {$1\times e^{-6}$}. The rollout numbers for the grounding ($G_1$) and answering ($G_2$) stages were set to 4 and 2, respectively. The threshold $\tau$ was set to 0.15. We used a default sampling temperature of 1 and a maximum response length of 512 tokens. All experiments were run on 8$\times$H100 GPUs, with a total training time of approximately 12 hours.

\subsection{Main Results}

\begin{table*}[t]
\belowrulesep=0pt
\aboverulesep=0pt
    \centering
    \renewcommand\arraystretch{1.1}
    \setlength{\tabcolsep}{1.5mm}
    \caption{\textbf{Performance comparisons on high-resolution reasoning benchmarks.} $^*$ denotes the results sourced from DeepEyes \cite{zheng2025deepeyes} and $^\dagger$ denotes the results sourced from Mini-o3 \cite{lai2025mini}. The \textbf{bold} values correspond to the best-performing results in each column, while the \underline{underlined} values correspond to the second-best.}
    %\resizebox{\textwidth}{!}{
    \scalebox{0.96}{
    \begin{tabular}{l c | c c c|c c c|c c c } 
    \toprule
    %\rowcolor{gray!10} 
    \multirow{2}{*}{\textbf{Model}}  & \multirow{2}{*}{\textbf{\makecell{Param \\ Size} }} &  \multicolumn{3}{c|}{\textbf{$V^*$} \textbf{Bench} \cite{wu2024v}} & \multicolumn{3}{c|}{\textbf{HR-Bench 4K} \cite{wang2025divide}} & \multicolumn{3}{c}{\textbf{HR-Bench 8K} \cite{wang2025divide}}\\
    &  & \textbf{Attr} & \textbf{Spatial} & \textbf{Overall} & \textbf{FSP} & \textbf{FCP} & \textbf{Overall} & \textbf{FSP} & \textbf{FCP} & \textbf{Overall}\\
    \midrule
    GPT-4o~\cite{achiam2023gpt}  & - & - & - &66.0 & 70.0 & 48.0 &   59.0  & 62.0  & 49.0 & 55.5 \\
    Gemini Pro \cite{reid2024gemini} & - &  40.9 & 59.2 & 48.2 & - & - & - & - & - & - \\
    GPT-4V \cite{gpt4v} & - & 51.3 & 60.5 & 55.0 & - & - & - & - & - & - \\
    Gemini 1.5 Flash \cite{reid2024gemini} & - & - & - & - & 76.8 & 56.8 & 66.8 & 69.2 & 56.7 & 62.8  \\
    \hdashline 
    %o3~\cite{o3}  & - & -  & - & 95.7 & - & - & - & - & - & - \\
    %\midrule
    SEAL~\cite{wu2024v}  & 7B & 74.8 & 76.3 & 75.4 & - & - & - & - & - & - \\
    DyFo~\cite{li2025dyfo} & 7B & 80.0 & 82.9 & 81.2 & - & - & - & - & - & - \\
    MGPO \cite{huang2025high} & 7B &  - & - &76.4 & - & - & - & - & - & -\\
    %ZoomEye~\cite{shen2024zoomeye} & \ding{55} & 7B & 93.9 & 85.5 & 90.6 & 84.3 & 55.0 & 69.6 & 88.5 & 50.0 & 69.3 \\
    %\midrule
    LLaVA-OneVision \cite{li2024llava} & 7B  & 75.7 & 75.0 & 75.4 & 72.0 & 54.0 & 63.0 & 67.3 & 52.3 & 59.8 \\
    Qwen2.5-VL$^{*}$ \cite{bai2025qwen2}  & 7B & 73.9 & 67.1 & 71.2 & 85.2 & 52.2 & 68.8 & 78.8 & 51.8 & 65.3 \\
    Qwen2.5-VL$^{*}$~\cite{bai2025qwen2}  & 32B & 87.8 & 88.1 & 87.9 & 89.8 & 58.0 & 73.9 & 84.5 & 56.3 & 70.4 \\
    DeepEyes$^{*}$ \cite{zheng2025deepeyes}  & 7B & \underline{91.3} & \underline{88.2} & 90.1 & \underline{91.3} & 59.0 & 75.1 & \underline{86.8} & 58.5 & 72.6 \\
    DeepEyes$^\dagger$ \cite{zheng2025deepeyes}& 7B & - & - & 83.3 & - &  - & 73.2 & - & - &  69.5 \\
    Mini-o3$^\dagger$ \cite{lai2025mini}  & 7B & - & - & 88.2 & - &  - & \underline{77.5} & - & - &  \underline{73.3} \\
    Chain-of-Focus$^\dagger$ \cite{zhang2025chain}  & 7B & - & - & 88.0 & - & - & - & - & - & - \\
    Pixel Reasoner$^\dagger$ \cite{wang2025pixel} & 7B & - & - & 86.3 & - & - & 74.0 & - & - & 66.9 \\
    TreeVGR \cite{bai2025multi} & 7B & \textbf{94.0} & 87.0 & \textbf{91.1} & 90.3 & \textbf{64.0} & 77.1 & 86.5 & \underline{59.8} & 73.1 \\
    \textbf{TiCO (Ours)} & 7B & \underline{91.3} & \textbf{89.5} & \underline{90.6} & \textbf{92.3} & \underline{63.8} & \textbf{78.0} & \textbf{91.0} & \textbf{61.5} & \textbf{76.3} \\
    %\midrule
    \bottomrule
\end{tabular}
    }
    \label{tab:resHighResolution}
\end{table*}

% -------------

\begin{table*}[t]
\belowrulesep=0pt
\aboverulesep=0pt
    \centering
    \renewcommand\arraystretch{1.1}
    \setlength{\tabcolsep}{2.5mm}
    \caption{\textbf{Performance comparisons on VisualProbe  and general reasoning benchmarks.} For VisualProbe$^{*}$ \cite{lai2025mini} and MME-RealWorld$^{*}$ (MME-RW) \cite{zhang2024mme}, the results of the compared methods are sourced from Mini-o3 \cite{lai2025mini}. For POPE$^{\dagger}$ \cite{li2023evaluating}, the results of the compared methods are sourced from DeepEyes \cite{zheng2025deepeyes}. The \textbf{bold} values correspond to the best-performing results in each column, while the \underline{underlined} values correspond to the second-best.}
    %\resizebox{\textwidth}{!}{
    \scalebox{0.96}{
    \begin{tabular}{l|ccc|c|cccccc} 
    \toprule
    %\rowcolor{gray!10} 
    \multirow{2}{*}{\textbf{Model}}  &  \multicolumn{3}{c|}{\textbf{VisualProbe}$^{*}$} & \multirow{2}{*}{\textbf{\makecell{MME\\-RW$^{*}$}}}  & \multicolumn{4}{c}{\textbf{POPE}$^{\dagger}$} \\
    & \textbf{hard} & \textbf{medium} & \textbf{easy} & & \textbf{Adv}  & \textbf{Pop}  &  \textbf{Random} &  \textbf{Overall} \\
    \midrule
    GPT-4o \cite{achiam2023gpt} & 11.2 & 15.4 & 47.5 & 45.2 & - & - &- & -     \\
    LLaVA-OneVision \cite{li2024llava} &  13.4 &  12.5 & 36.2 & 57.4 & - &- & - & \underline{88.4} \\
    Qwen2.5-VL \cite{bai2025qwen2}  & 23.9 & 26.0 & 39.1 & 57.3 & \underline{85.9} & 86.5 & 87.2 & 85.9 \\
    Pixel Reasoner \cite{wang2025pixel} & 28.8 & 29.6 & 58.4 & 64.4 & -  &  - &  - & - \\
    DeepEyes \cite{zheng2025deepeyes} & 35.1 & 29.8 & 60.1 &  64.0  & 84.0 & \underline{87.5} & \textbf{91.8} & 87.7 \\
    Mini-o3 \cite{lai2025mini} & \underline{48.0} & \underline{50.4} & \underline{67.0} & \underline{65.5} & - & - &- & -  \\
    \textbf{TiCO (Ours)} & \textbf{51.9} & \textbf{51.1} & \textbf{68.8} & \textbf{66.9} & \textbf{87.1} & \textbf{88.9} & \underline{91.4} & \textbf{89.1} \\
    \bottomrule
    \end{tabular}
    }
    \label{tab:resGeneral}
\end{table*}

\noindent \textbf{Comparisons on High-Resolution Reasoning Benchmarks.} Table \ref{tab:resHighResolution} and Table \ref{tab:resGeneral} (left) present performance comparisons on high-resolution reasoning benchmarks, including V* Bench \cite{wu2024v}, HR-Bench \cite{wang2025divide} at 4K/8K resolutions, and VisualProbe \cite{lai2025mini}. In general, our TiCO demonstrates strong and consistent performance advantages on high-resolution settings, particularly on HR-Bench. On HR-Bench 4K, TiCO achieves an overall accuracy of 78.0\%, outperforming the officially reported results of Mini-o3 (77.5\%), DeepEyes (75.1\%), and TreeVGR \cite{bai2025multi} (77.1\%). For both Fine-grained Single-instance Perception (FSP) and Fine-grained Cross-instance Perception (FCP), our method consistently achieves superior performance. This demonstrates its effectiveness in capturing subtle intra-object details as well as modeling complex inter-object relationships. In addition, the advantage becomes more pronounced on HR-Bench 8K, where TiCO reaches an overall accuracy of 76.3\%, clearly surpassing Mini-o3 (73.3\%) and DeepEyes (72.6\%). In addition, we evaluate our TiCO on VisualProbe benchmark \cite{lai2025mini}, which necessitates significantly longer interaction chains and more complex multi-turn reasoning. As shown in Table \ref{tab:resGeneral} (left), TiCO achieves state-of-the-art performance across all difficulty levels, notably outperforming Mini-o3 by 3.9\% on the "hard" split. These results indicate that the coordinate-free evidence rollout mechanism is particularly effective in high-resolution scenarios, where fine-grained evidence localization is more challenging and critical. % These results underscore the robustness of our coordinate-free paradigm in long-horizon scenarios.

On V* Bench \cite{wu2024v}, TiCO remains competitive but does not achieve the best overall score compared to TreeVGR \cite{bai2025multi} in Table \ref{tab:resHighResolution}. While TiCO achieves the best spatial performance, it shows slightly lower attribute and overall accuracy. This discrepancy can be attributed to the differing nature of the two subsets: attribute recognition depends on fine-grained, instance-specific appearance cues (\eg, color or material), whereas spatial reasoning emphasizes relational understanding across multiple regions. As TiCO prioritizes semantic region interaction and distribution-aware evidence aggregation, it is naturally better aligned with relational reasoning. In contrast, attribute-level tasks are more sensitive to fine-grained local textures, which may be partially filtered out during the semantic evidence aggregation process intended for high-level reasoning.

%In contrast, attribute-level tasks are more sensitive to precise local details, which may incur minor performance trade-offs.

%One possible explanation is that V* Bench places stronger emphasis on explicit spatial relationships and global geometric reasoning. Although TiCO effectively selects semantically salient regions via similarity-based connected components, its evidence sampling mechanism may prioritize high-response local patches while underrepresenting auxiliary contextual regions necessary for modeling precise relational constraints. 

\noindent \textbf{Comparisons on General Reasoning Benchmarks.} Following \cite{lai2025mini,zheng2025deepeyes}, we further evaluate TiCO on MME-RealWorld (MME-RW) \cite{zhang2024mme} and POPE \cite{li2023evaluating}, two representative benchmarks for assessing general visual reasoning. MME-RW emphasizes real-world, high-resolution images with complex scenes, requiring fine-grained perception and grounded reasoning. In contrast, POPE targets object hallucination by evaluating whether model responses are faithfully grounded in visual evidence rather than language priors.

As summarized in Table \ref{tab:resGeneral}, our TiCO achieves competitive performance on both benchmarks, with its leading score of 66.9\% on MME-RW and an overall score of 89.1\% on POPE. Compared to the baseline Qwen2.5-VL, TiCO exhibits a substantial performance leap, particularly on MME-RW (66.9\% vs. 57.3\%), suggesting that our interleaved reasoning paradigm overcomes the perceptual bottlenecks inherent in standard text-centric reasoning. While the current leading method Mini-o3 shows strong capabilities, TiCO surpasses it by 1.4\% on MME-RW, primarily because our coordinate-free anchoring avoids the information loss associated with the rigid bounding box regression. 

\begin{table*}[t]
\belowrulesep=0pt
\aboverulesep=0pt
    \centering
    \renewcommand\arraystretch{1.1}
    \setlength{\tabcolsep}{3mm}
    \caption{\textbf{Efficiency analysis.} ``Overall" denotes the overall accuracy. ``\#Token" denotes the average number of tokens contained in the generated answers. ``Time" denotes the average inference time per query (in seconds).}
    %\resizebox{\textwidth}{!}{
    \scalebox{0.96}{
    \begin{tabular}{l |c c c |c c c} 
    \toprule
    %\rowcolor{gray!10} 
    \multirow{2}{*}{\textbf{Model}}   & \multicolumn{3}{c|}{\textbf{HR-Bench 4K} \cite{wang2025divide}} & \multicolumn{3}{c}{\textbf{HR-Bench 8K} \cite{wang2025divide}}\\
    & \textbf{Overall}$\uparrow$ & \textbf{\#Token}$\downarrow$ & \textbf{Time}$\downarrow$ & \textbf{Overall}$\uparrow$ & \textbf{\#Token}$\downarrow$ & \textbf{Time}$\downarrow$ \\
    \midrule
    Mini-o3 \cite{lai2025mini}  & 77.5 & 527.7 & 20.2 & 73.3 & 714.5 & 35.1  \\
    DeepEyes \cite{zheng2025deepeyes} & 73.2 & 398.2 & 18.3 & 69.5 & 419.3 & 27.4 \\
    \textbf{TiCO (Ours)}  & \textbf{78.0} & \textbf{6.2} & \textbf{1.9} & \textbf{76.3} & \textbf{6.1} & \textbf{3.6} \\
    %\midrule
    \bottomrule
\end{tabular}
    }
    \label{tab:efficiency}
\end{table*}

\noindent \textbf{Efficiency Analysis.} To evaluate inference efficiency, we introduce two additional metrics beyond overall accuracy: \emph{\#Token}, the average number of generated tokens per query, and \emph{Time}, the average inference time (in seconds). These metrics reflect the computational overhead of long-form reasoning. In Table \ref{tab:efficiency}, we conduct an efficiency analysis comparing two well-known coordinate-based methods, Mini-o3 \cite{lai2025mini} and DeepEyes \cite{zheng2025deepeyes}. For a fair comparison, we directly use their released checkpoints without employing inference service engines vLLM \cite{kwon2023efficient}, in order to better evaluate their inherent efficiency rather than performance boosted by system-level acceleration. As shown in Table \ref{tab:efficiency}, our TiCO achieves higher accuracy while using only a tiny fraction of the computational budget. On HR-Bench 4K, TiCO attains 78.0\% overall accuracy using only 6.2 tokens, which is about 1.2\% of Mini-o3's 527.7 tokens and 1.6\% of DeepEyes' 398.2 tokens. Meanwhile, the inference time is reduced to 1.9s, roughly 9.4\% of Mini-o3 (20.2s) and 10.4\% of DeepEyes (18.3s). These results demonstrate that TiCO delivers superior reasoning performance while maintaining high efficiency in terms of response tokens and inference time, highlighting the significant advantages of our coordinate-free approach over existing coordinate-based methods.

\subsection{Ablations on Architectural Design}

\noindent \textbf{Coordinate-based \vs Coordinate-free Reasoning.} To validate our proposed coordinate-free design, we first compare TiCO with the coordinate-based reasoning variants under identical backbones and training settings. Specifically, we evaluate against two pipelines: 1) \emph{Coordinate-based with thinking}: The model first generates intermediate reasoning tokens (\texttt{<thinking>}...\texttt{</thinking>}), followed by explicit bounding-box prediction (\texttt{<box>}...\texttt{</box>}), and then produces the final answer (\texttt{<ans>}...\texttt{</ans>}); 2) \emph{Coordinate-based without thinking}: The model directly predicts a bounding box and subsequently generates answers conditioned on the cropped region without explicit intermediate reasoning tokens. 

As shown in Table \ref{tab:ablationArchi} (top), our coordinate-free approach consistently outperforms both coordinate-based variants across all benchmarks. This demonstrates that the coordinate-free approach effectively utilizes latent semantic anchors to provide a more flexible, high-fidelity representation of visual evidence. Interestingly, on certain benchmarks, the coordinate-based \emph{without} thinking variant slightly outperforms the \emph{with} thinking version. For example, on V* Bench \cite{wu2024v}, it achieves higher spatial (82.9\% vs. 80.3\%) and overall scores (84.3\% vs. 83.2\%), and on HR-Bench 4K \cite{wang2025divide}, it obtains a slightly better FCP score (62.5\% vs. 62.0\%). We attribute this to the error-sensitive nature of coordinate-based grounding, where inaccuracies in bounding-box localization can be amplified by additional reasoning tokens, since subsequent inference is strictly conditioned on discrete and potentially misaligned predictions. Refer to the supplementary material for more details.

\begin{table*}[t]
\belowrulesep=0pt
\aboverulesep=0pt
    \centering
    \renewcommand\arraystretch{1.1}
    \setlength{\tabcolsep}{1.5mm}
    \caption{\textbf{Ablation studies on the architectural design.} \textbf{Top:} Comparison between our proposed coordinate-free approach and coordinate-based counterparts; \textbf{Middle:} Comparison of different probing mechanisms, specifically probing via text embedding versus  probing via \texttt{<GROUND>} token. \textbf{Bottom:} Comparison of visual input configurations, including evidence-only, full-image-only, and their integrated setting.}
    %\resizebox{\textwidth}{!}{
    \scalebox{0.96}{
    \begin{tabular}{l | c c c|c c c|c c c } 
    \toprule
    %\rowcolor{gray!10} 
     \multirow{2}{*}{\textbf{Method}}  &  \multicolumn{3}{c|}{\textbf{$V^*$} \textbf{Bench} \cite{wu2024v}} & \multicolumn{3}{c|}{\textbf{HR-Bench 4K} \cite{wang2025divide}} & \multicolumn{3}{c}{\textbf{HR-Bench 8K} \cite{wang2025divide}}\\
    &  \textbf{Attr} & \textbf{Spatial} & \textbf{Overall} & \textbf{FSP} & \textbf{FCP} & \textbf{Overall} & \textbf{FSP} & \textbf{FCP} & \textbf{Overall}\\
    \midrule
    %(a) & \textbf{TiCO (Ours)} & \textbf{91.3} & 85.5 & \underline{89.0} & \textbf{92.3} & \textbf{63.8} & \textbf{78.0} & \textbf{91.0} & \textbf{61.5} & \textbf{76.3} \\
    %\hdashline
    {Coord-based \emph{w/} thinking} & 85.2 & 80.3 & 83.2 & 90.5 & 62.0 & 76.3 & 88.0 & 53.8 & 73.6 \\
    {Coord-based \emph{w/o} thinking} & 85.2 & 82.9 & 84.3 & 89.8 & 62.5 & 76.1 & 86.5 & 58.0 & 72.3 \\
    \textbf{Coord-free (Ours)} & \textbf{91.3} & \textbf{89.5} & \textbf{90.6} & \textbf{92.3} & \textbf{63.8} & \textbf{78.0} & \textbf{91.0} & \textbf{61.5} & \textbf{76.3}  \\
    \hdashline
    Probe via text embedding & 88.7 & 82.9 & 86.4 & 91.0 & 63.5 & 77.3 & 88.0 & 58.3 & 73.1\\
    \textbf{Probe via \texttt{<GROUND>} token} & \textbf{91.3} & \textbf{89.5} & \textbf{90.6} & \textbf{92.3} & \textbf{63.8} & \textbf{78.0} & \textbf{91.0} & \textbf{61.5} & \textbf{76.3} \\
    \hdashline
    Evidence only & 87.8 & 75.0 & 82.7 & 90.8 & 47.3 & 69.0 & 87.3 & 48.5 & 67.9 \\
    Full image only & 82.6 & 77.6 & 80.6 & 86.3 & 60.0 & 73.1 & 73.3 & 56.5 & 64.9\\
    \textbf{Full image + Evidence} & \textbf{91.3} & \textbf{89.5} & \textbf{90.6} & \textbf{92.3} & \textbf{63.8} & \textbf{78.0} & \textbf{91.0} & \textbf{61.5} & \textbf{76.3} \\
    %\midrule
    \bottomrule
\end{tabular}
    }
    \label{tab:ablationArchi}
\end{table*}

\noindent \textbf{Ablation Studies on \texttt{<GROUND>} Token.} To evaluate the necessity of the proposed \texttt{<GROUND>} token as a dedicated semantic probe, we compare TiCO with a variant that replaces \texttt{<GROUND>} by pooled textual embeddings for patch selection. Specifically, we compute a global textual representation by mean-pooling the query token embeddings and directly use it to estimate the cosine similarity distribution over visual patches.  

As shown in the middle part of Table \ref{tab:ablationArchi}, probing via \texttt{<GROUND>} token consistently outperforms probing via text embedding across all metrics. This performance gain suggests that a dedicated probe token is better equipped to bridge the modality gap during complex reasoning than text embeddings.

\noindent \textbf{Do evidence regions contribute to reasoning?} To evaluate the necessity of the thinking-with-image paradigm, we compare three visual input configurations: (1) \emph{Full image only}, which performs vanilla reasoning directly on the input image; (2) \emph{Evidence only}, which removes the full-image context from the reasoning history and generates answers solely based on the extracted evidence regions; and (3) \emph{Full image + Evidence}, corresponding to our proposed thinking-with-image paradigm that integrates both global context and localized evidence. 

As shown in Table \ref{tab:ablationArchi} (Bottom), the full-image-only setting yields the weakest overall performance (\eg, 64.9\% on HR-Bench 8K), indicating that holistic reasoning without explicit grounding is insufficient for high-resolution perception. The evidence-only variant improves over full-image-only in several cases (\eg, 82.7\% \vs 80.6\% on V$^{*}$ Bench), demonstrating that focused regional evidence provides more task-relevant cues. These results confirm that the thinking-with-image paradigm effectively unifies coarse-to-fine visual reasoning, leading to more accurate and reliable predictions.

\begin{figure*}[t]
	\centering
    \includegraphics[width=0.86\textwidth]{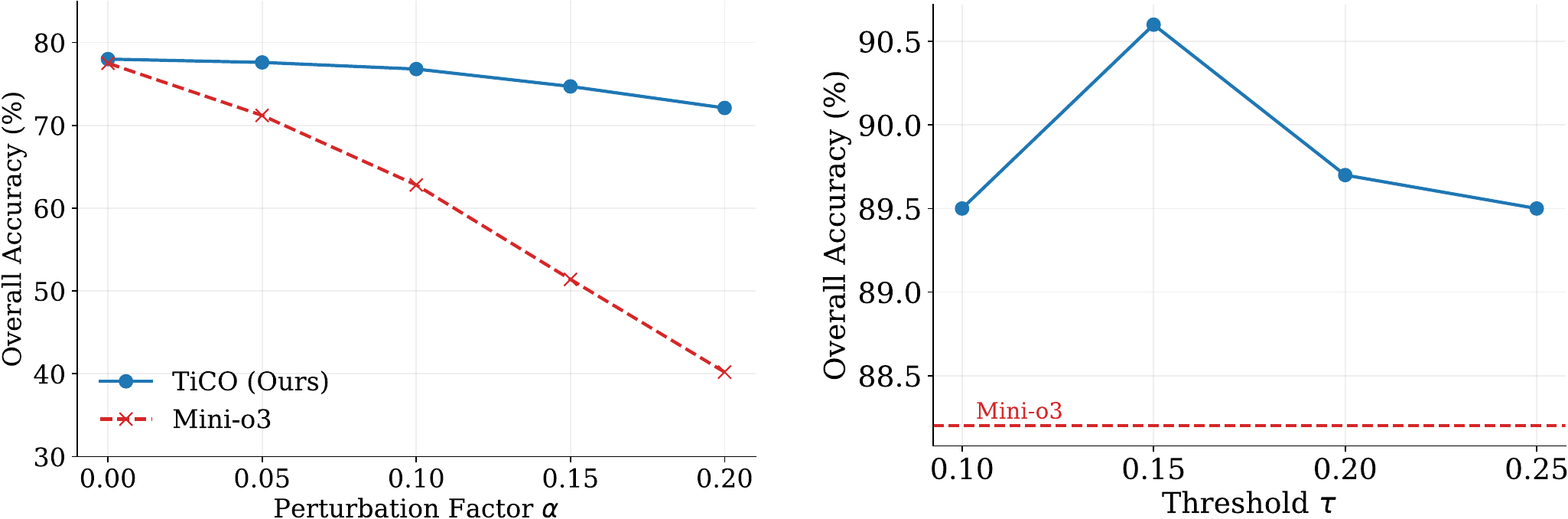}
    %\vspace{-1mm}
    \caption{\textbf{Left: Robustness analysis} on HR-Bench 4K \cite{wang2025divide}; \textbf{Right: Hyper-parameter analysis} of threshold $\tau$ on V$^{*}$ benchmark \cite{wu2024v}.}
	\label{fig:ablateTau}
	%\vspace{-2mm}
\end{figure*}

% \begin{figure}[t]
%     \centering
%     \begin{minipage}[t]{0.45\linewidth}
%         \centering
%         \includegraphics[width=\linewidth]{figs/robust.pdf}
%         \caption{\textbf{Robustness analysis} on HR-Bench 4K.}
%         \label{fig:ablateAlpha}
%     \end{minipage}
%     \hspace{5mm}
%     \begin{minipage}[t]{0.45\linewidth}
%         \centering
%         \includegraphics[width=\linewidth]{figs/threshold.pdf}
%         \caption{\textbf{Hyper-parameter analysis} of threshold $\tau$ on V$^{*}$ benchmark.}
%         \label{fig:ablateTau}
%     \end{minipage}
% \end{figure}

\noindent \textbf{Robustness Analysis.} To assess the robustness of coordinate-based (\eg, Mini-o3) and coordinate-free methods, we conduct experiments under \emph{noisy} bounding-box supervision. Specifically, we introduce controlled noise to the bounding boxes by applying random spatial perturbations, where the perturbation factor $\alpha$ denotes the relative magnitude of translation and scale jitter with respect to the box size. Refer to the supplementary material for more details.

As shown in Fig. \ref{fig:ablateTau} (left), we observe that Mini-o3 degrades rapidly as the perturbation factor $\alpha$ increases, whereas TiCO maintains relatively stable performance even under large spatial perturbations. This performance gap stems from the fundamental difference in supervision and reasoning paradigms. In coordinate-based models, noisy bounding boxes directly corrupt the numerical targets, causing the regression head to output entirely misaligned coordinates that fail to capture the target object. In contrast, TiCO is supervised by dense patch-wise labels, where the inclusion of spatial noise only affects the binary labels of a small subset of boundary patches. Since the labels for the remaining core patches stay correct, the semantic probe token can still reliably identify the underlying connected components and sample the necessary visual evidence. These results demonstrate that our coordinate-free design provides substantially improved robustness to imperfect bounding-box annotations.

% \lipsum[1] \textcolor{red}{Figure} \ref{fig:ablateTau} (left) \textcolor{red}{TODO}

\subsection{Ablations on Evidence Region Extraction}

\noindent \textbf{Ablation studies on evidence region extraction.} As described in Section \ref{sec:3.2}, our method adopts \emph{importance sampling} over connected components, where each candidate region is sampled according to its aggregated similarity score. This distribution-driven strategy prioritizes semantically confident regions while preserving rollout diversity. To assess the necessity of this design, we compare it against three alternative evidence selection schemes under identical training and rollout settings: 1) \emph{Random sampling}: uniformly sampling connected components without considering similarity scores; 2) \emph{Deterministic top-scoring sampling}: A set of $N_1$ highest-scoring regions is deterministically selected; 3) \emph{Token-wise sampling}: Individual high-confidence patches are sampled directly from the similarity distribution without grouping them into connected regions.

As demonstrated in Table \ref{tab:ablationEvidence} (top), our proposed importance sampling consistently achieves the best performance across all benchmarks, notably reaching 90.6\% on $V^*$ Bench and 76.3\% on HR-Bench 8K. Although top-scoring sampling ranks as the second-best strategy by selecting regions with the highest deterministic scores, its inherent rigidity limits exposure to diverse visual cues and risks omitting critical, lower-scoring secondary evidence. In contrast, the token-wise sampling performs the worst, with scores dropping to 81.2\% on $V^*$ Bench and 65.0\% on HR-Bench 8K. This decline primarily stems from the lack of regional grouping, where selecting independent patches without considering connected components leads to semantically fragmented evidence that hinders the model's ability to reconstruct coherent object-level information. 
%These results validate that our importance sampling effectively strikes a balance—maintaining the semantic integrity of grouped regions while utilizing distribution-driven selection to preserve the regional diversity necessary for complex reasoning.

\begin{table*}[t]
\belowrulesep=0pt
\aboverulesep=0pt
    \centering
    \renewcommand\arraystretch{1.1}
    \setlength{\tabcolsep}{1.5mm}
    %\caption{\textbf{Ablation studies on evidence region extraction.} \textbf{Top: Comparison of sampling strategies} including Top-scoring (deterministic highest-score selection), random (uniform selection), token-wise (independent patch selection without grouping), and our importance sampling; \textbf{Bottom: Hyper-parameter analysis} including the number of evidence region $N_1$ and answer rollout $N_2$. The \textbf{bold} values correspond to the best-performing results, while the \underline{underlined} values correspond to the second-best.}
    \caption{\textbf{Comparisons of sampling strategies} including top-scoring (deterministic highest-score selection), random (uniform selection), token-wise (independent patch selection without grouping), and our importance sampling. The \textbf{bold} values denote the best-performing results, while the \underline{underlined} values denote the second-best results.}
    %\resizebox{\textwidth}{!}{
    \scalebox{0.96}{
    \begin{tabular}{l | c c c|c c c|c c c } 
    \toprule
    %\rowcolor{gray!10} 
    \multirow{2}{*}{\textbf{Method}}  &  \multicolumn{3}{c|}{\textbf{$V^*$} \textbf{Bench} \cite{wu2024v}} & \multicolumn{3}{c|}{\textbf{HR-Bench 4K} \cite{wang2025divide}} & \multicolumn{3}{c}{\textbf{HR-Bench 8K} \cite{wang2025divide}}\\
    &  \textbf{Attr} & \textbf{Spatial} & \textbf{Overall} & \textbf{FSP} & \textbf{FCP} & \textbf{Overall} & \textbf{FSP} & \textbf{FCP} & \textbf{Overall}\\
    \midrule
    Top-scoring sampling & \underline{88.7} & \underline{80.3} & \underline{85.3} & 91.5 & \underline{61.5} & 76.5 & \underline{87.5} & 59.5 & \underline{73.5}\\
    Random sampling & 87.8 & 79.0 & 84.3 & \textbf{93.0} & 60.8 & \underline{76.9} & 86.3 & \underline{60.0} & 73.1\\
    Token-wise sampling & 81.7 & \underline{80.3} & 81.2 & 87.5 & 59.0 & 73.3 & 76.3 & 53.8 & 65.0\\
    \textbf{Importance sampling} & \textbf{91.3} & \textbf{89.5} & \textbf{90.6} & \underline{92.3} & \textbf{63.8} & \textbf{78.0} & \textbf{91.0} & \textbf{61.5} & \textbf{76.3} \\
    % \hdashline
    % $N_1=1$, $N_2=8$ & 87.8 & 80.3 & 84.8 & 90.8 & 62.3 & 76.5 & 88.5 & 58.5 & 73.5 \\
    % $N_1=2$, $N_2=4$ & \underline{89.6} & \textbf{86.8} & \underline{88.5} & \textbf{92.5} & \underline{62.8} & \underline{77.6} & \underline{89.8} & \underline{58.8} & \underline{74.3}  \\
    % $N_1=4$, $N_2=2$ & \textbf{91.3} & \textbf{86.8} & \textbf{89.5} & \underline{92.3} & \textbf{63.8} & \textbf{78.0} & \textbf{91.0} & \textbf{61.5} & \textbf{76.3} \\
    \bottomrule
\end{tabular}
    }
    \label{tab:ablationEvidence}
\end{table*}

\begin{table*}[t]
\belowrulesep=0pt
\aboverulesep=0pt
    \centering
    \renewcommand\arraystretch{1.1}
    \setlength{\tabcolsep}{1.5mm}
    \caption{\textbf{Ablations on training configurations} including SFT and RL stages. "SFT (\emph{w/o} ground loss)" denotes removing the grounding loss (Equ \eqref{eq:5}) in SFT.}
    %\resizebox{\textwidth}{!}{
    \scalebox{0.96}{
    \begin{tabular}{l | c c c|c c c|c c c } 
    \toprule
    %\rowcolor{gray!10} 
     \multirow{2}{*}{\textbf{Method}}  &  \multicolumn{3}{c|}{\textbf{$V^*$} \textbf{Bench} \cite{wu2024v}} & \multicolumn{3}{c|}{\textbf{HR-Bench 4K} \cite{wang2025divide}} & \multicolumn{3}{c}{\textbf{HR-Bench 8K} \cite{wang2025divide}}\\
    &  \textbf{Attr} & \textbf{Spatial} & \textbf{Overall} & \textbf{FSP} & \textbf{FCP} & \textbf{Overall} & \textbf{FSP} & \textbf{FCP} & \textbf{Overall}\\
    \midrule
    %\hdashline
    SFT Only & 87.0 & 81.6 & 84.8 & 90.8 & 61.5 & 76.1 & 86.5 & 55.8 & 71.1\\
    RL Only & 79.1 & 73.7 & 77.0 & 85.8 & 54.3 & 70.0 & 82.0 & 55.8 & 68.9 \\
    SFT (\emph{w/o} ground loss) + RL & 77.4 & 72.4 & 75.4 & 77.8 & 59.3 & 68.5 & 66.3 & 58.8 & 62.5\\
    \textbf{SFT + RL (Ours)} & \textbf{91.3} & \textbf{89.5} & \textbf{90.6} & \textbf{92.3} & \textbf{63.8} & \textbf{78.0} & \textbf{91.0} & \textbf{61.5} & \textbf{76.3} \\
    \bottomrule
\end{tabular}
    }
    \label{tab:ablationTrain}
\end{table*}

\noindent \textbf{Hyper-parameter analysis.} To evaluate the robustness of our importance sampling mechanism, we conduct a sensitivity analysis on the selection threshold $\tau$ in Equ \eqref{eq:2}. We vary $\tau$ from 0.10 to 0.25 and report the overall accuracy on $V^*$ benchmark. For reference, we also include the performance of the coordinate-based baseline Mini-o3 as a dashed line in Figure \ref{fig:ablateTau} (right). As shown, TiCO consistently outperforms  Mini-o3 across all threshold values, demonstrating robustness to threshold selection. %We adopt $\tau=0.15$, as it achieves consistently strong results across three benchmarks.

% We evaluate the trade-off between the number of evidence regions $N_1$ and answer rollouts $N_2$ under a fixed budget ($N_1 \times N_2 = 8$). As shown in Table \ref{tab:ablationEvidence} (bottom), the configuration ($N_1=4, N_2=2$) achieves the best overall performance, suggesting that allocating more capacity to evidence coverage is more beneficial than increasing rollout multiplicity. Allocating more evidence regions improves coverage and grounding reliability, while a moderate number of rollouts remains sufficient for stable answer generation.

\subsection{Ablations on Training Strategy}

\noindent \textbf{Ablation of SFT and RL stages.} We conduct the ablation studies on the SFT and RL training stages. As shown in Table \ref{tab:ablationTrain}, our joint training paradigm (SFT + RL) yields the best performance across all metrics, significantly surpassing standalone training stages. This demonstrates that RL further refines the policy on top of supervised grounding and improves final prediction quality.

\noindent \textbf{Effect of dense supervision in SFT.} We further examine the role of dense grounding supervision (Equ \eqref{eq:6}) by comparing our full model with the "SFT ($w/o$ ground loss) + RL" variant. As shown in Table \ref{tab:ablationTrain}, removing the grounding loss results in substantial performance degradation (\eg, 75.4\% \vs 90.6\% on V$^{*}$ Bench and 62.5\% \vs 76.3\% on HR-Bench 8K), highlighting the critical role of dense spatial supervision. These results indicate that explicit grounding signals during SFT are essential for learning accurate evidence localization, which subsequently benefits RL fine-tuning and overall reasoning performance.

\begin{figure*}[t]
	\centering
    \includegraphics[width=\textwidth]{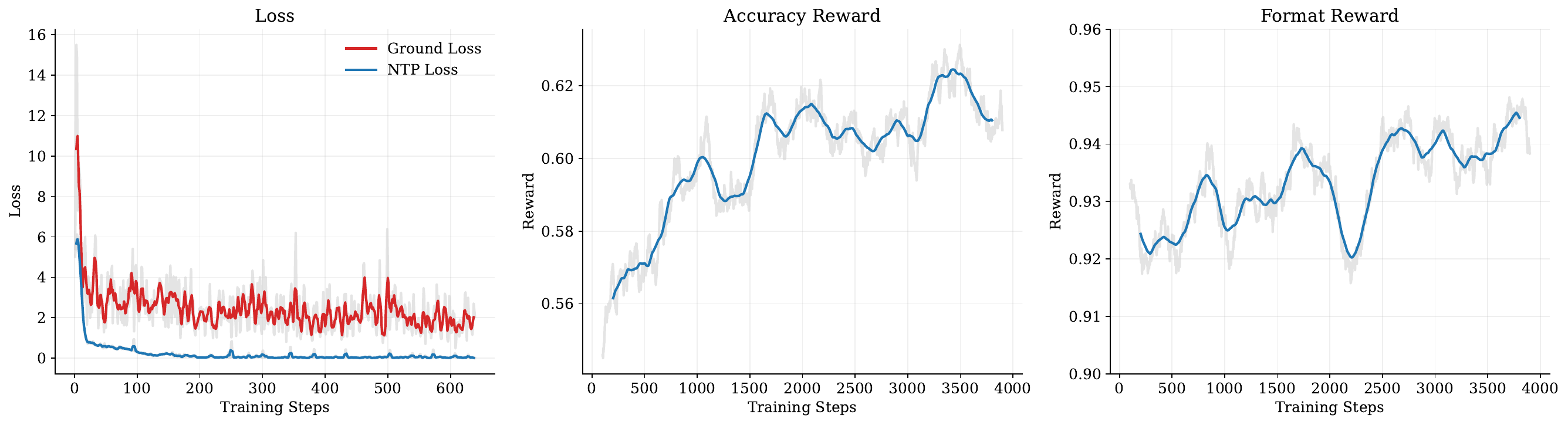}
    % \vspace{-1mm}
	\caption{\textbf{Training dynamics of TiCO during SFT and RL stages.} Left: The convergence of the SFT phase including the grounding loss and the next-token prediction (NTP) loss; Right: The accuracy reward and format reward over training steps.}
	\label{fig:curve}
	%\vspace{-2mm}
\end{figure*}

\noindent \textbf{Visualization of training dynamics.} As shown in Fig. \ref{fig:curve}, TiCO exhibits stable and well-behaved optimization dynamics across both training stages. During the SFT phase, both the grounding loss and the next-token prediction loss decrease rapidly in the early steps and gradually converge, indicating effective alignment between semantic probe supervision and language modeling objectives. In the subsequent RL phase, both the accuracy reward and the format reward steadily improve over training steps, demonstrating that the RL stage successfully refines both visual reasoning and output formatting without destabilizing training.

%These results confirm the effectiveness and stability of our two-stage optimization strategy.

% \subsection{Visualization}

% training dynamic (SFT-Two Losses; RL-Two Rewards)

% attention map (after SFT/RL)

% Main results update 

% Failure Case (multiple objects, spatial relationships)

% LIMITATIONS

% What if the selected regions incorrect?

% Supplementary:
% our SFT construction details

% Anonymous Github

% Prompts

% thinking vs non-thinking (1. Inconsistent; 2. Replication)

% Mini-o3 performance under controlled budget

% Robust Analysis (Define alpha)
\section{Conclusion}

In this paper, we revisit the prevailing coordinate-based paradigm in thinking-with-images frameworks and identify its inherent semantic and resolution gaps. To overcome these limitations, we introduce TiCO, a coordinate-free visual reasoning framework that shifts grounding from explicit geometric regression to semantic evidence discovery in the latent token space. By leveraging a dedicated probe token to estimate similarity distributions and performing importance sampling over spatially connected components, TiCO identifies semantically dominant evidence regions without the need for explicit coordinate prediction. Extensive experiments demonstrate that our TiCO not only improves reasoning accuracy but also significantly enhances inference efficiency.

\bibliographystyle{assets/plainnat}
\bibliography{resources/main}
\clearpage
\beginappendix

This supplementary material is organized as follows. We begin with a detailed description of more experimental results, then present additional experimental setups, followed by the visualization results.

Specifically, the experiment results include the following aspects:
\begin{itemize}[topsep=0pt, partopsep=0pt, leftmargin=28pt, parsep=0pt, itemsep=0pt]
    \item Ablation studies on threshold $\tau$.
    \item Hyper-parameter analysis.
    \item Performance under controlled budget.
\end{itemize}
The experimental setups include the following aspects:
\begin{itemize}[topsep=0pt, partopsep=0pt, leftmargin=28pt, parsep=0pt, itemsep=0pt]
    \item Construction details of SFT dataset.
    \item Evolution of attention maps from SFT to RL.
    \item Robustness Analysis.
    \item Failure examples of the "thinking" mode in coordinate-based methods.
    \item Prompt.
    \item Details of connected component generation.
    \item Limitations.
\end{itemize}
The visualization results include the following aspects:
\begin{itemize}[topsep=0pt, partopsep=0pt, leftmargin=28pt, parsep=0pt, itemsep=0pt]
    \item Failure cases of our TiCO.
    \item Comparison between coordinate-based methods and TiCO.
\end{itemize}

%\textcolor{blue}{\textbf{To facilitate reproducibility and further research, we provide the source code in the supplementary materials.}}

\section{Experimental Results}

\begin{table*}[t]
\belowrulesep=0pt
\aboverulesep=0pt
    \centering
    \renewcommand\arraystretch{1.1}
    \setlength{\tabcolsep}{1.5mm}
    \caption{\textbf{Hyper-parameter analysis} including the number of evidence region $N_1$ and answer rollout $N_2$. The \textbf{bold} values correspond to the best-performing results, while the \underline{underlined} values correspond to the second-best.}
    %\resizebox{\textwidth}{!}{
    \scalebox{0.96}{
    \begin{tabular}{l | c c c|c c c|c c c } 
    \toprule
    %\rowcolor{gray!10} 
    \multirow{2}{*}{\textbf{Method}}  &  \multicolumn{3}{c|}{\textbf{$V^*$} \textbf{Bench} \cite{wu2024v}} & \multicolumn{3}{c|}{\textbf{HR-Bench 4K} \cite{wang2025divide}} & \multicolumn{3}{c}{\textbf{HR-Bench 8K} \cite{wang2025divide}}\\
    &  \textbf{Attr} & \textbf{Spatial} & \textbf{Overall} & \textbf{FSP} & \textbf{FCP} & \textbf{Overall} & \textbf{FSP} & \textbf{FCP} & \textbf{Overall}\\
    \midrule
    $N_1=1$, $N_2=8$ & 87.8 & 80.3 & 84.8 & 90.8 & 62.3 & 76.5 & 88.5 & 58.5 & 73.5 \\
    $N_1=2$, $N_2=4$ & \underline{89.6} & \textbf{86.8} & \underline{88.5} & \textbf{92.5} & \underline{62.8} & \underline{77.6} & \underline{89.8} & \underline{58.8} & \underline{74.3}  \\
    $N_1=4$, $N_2=2$ & \textbf{91.3} & \textbf{86.8} & \textbf{89.5} & \underline{92.3} & \textbf{63.8} & \textbf{78.0} & \textbf{91.0} & \textbf{61.5} & \textbf{76.3} \\
    \bottomrule
\end{tabular}
    }
    \label{tab:ablationN}
\end{table*}

\begin{figure*}[b]
	\centering
    \includegraphics[width=\textwidth]{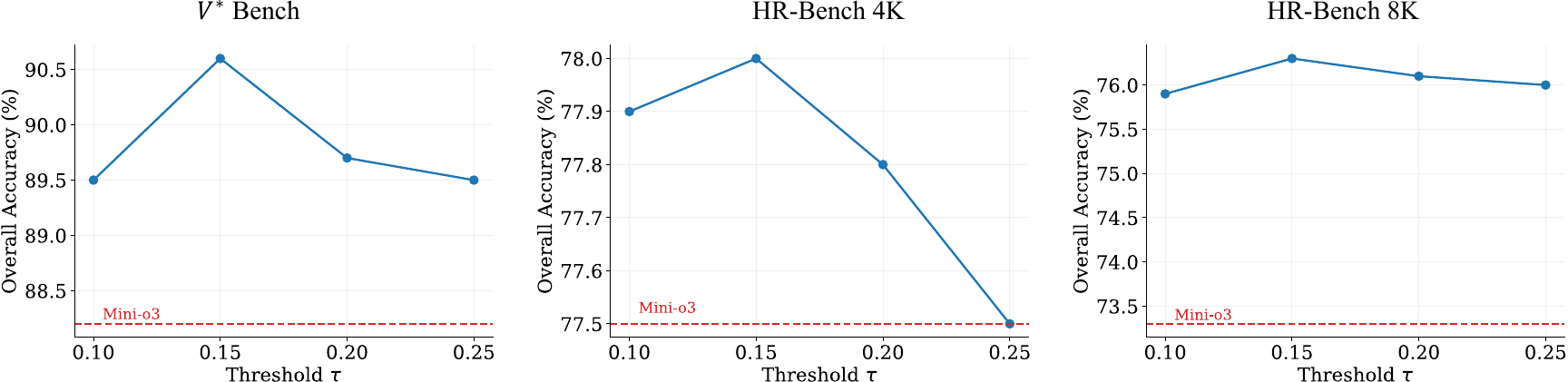}
    %\vspace{-1mm}
    \caption{\textbf{Hyper-parameter analysis} of threshold $\tau$.}
	\label{fig:ablateTauAll}
	%\vspace{-2mm}
\end{figure*}

\begin{figure}[t]
    \centering
    \begin{subfigure}[t]{0.49\linewidth}
        \centering
        \includegraphics[width=0.98\linewidth]{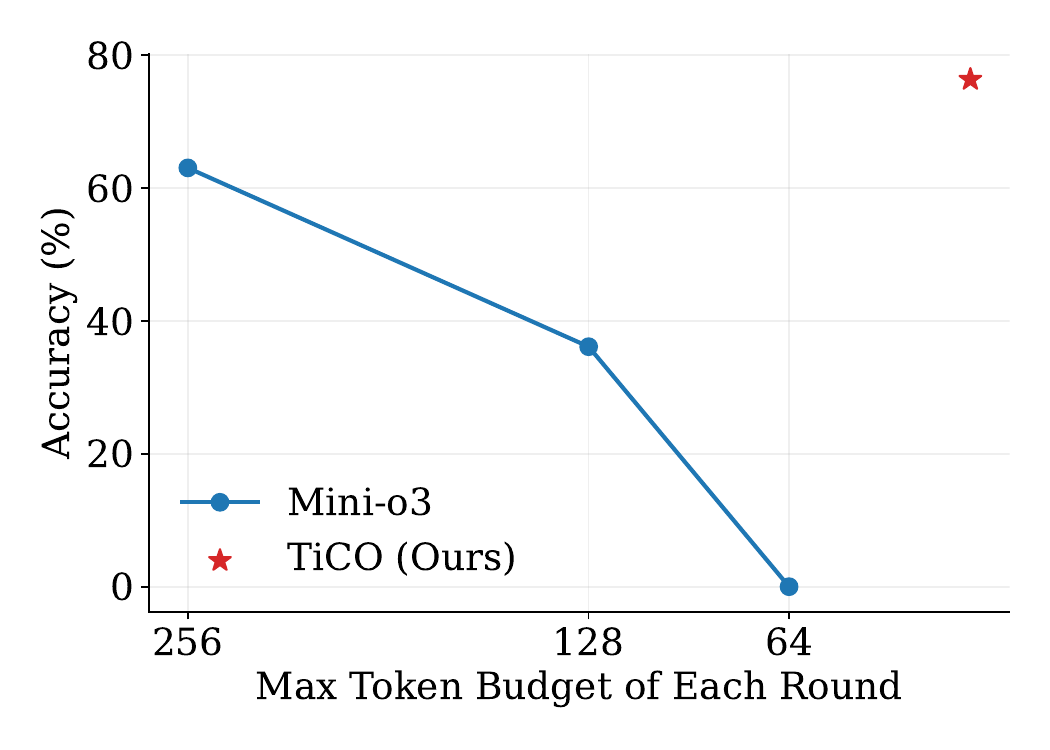}
        \caption{}
        \label{fig:budget}
    \end{subfigure}
    %\hfill
    %\vspace{1mm}
    \begin{subfigure}[t]{0.49\linewidth}
        \centering
        \includegraphics[width=0.6\linewidth]{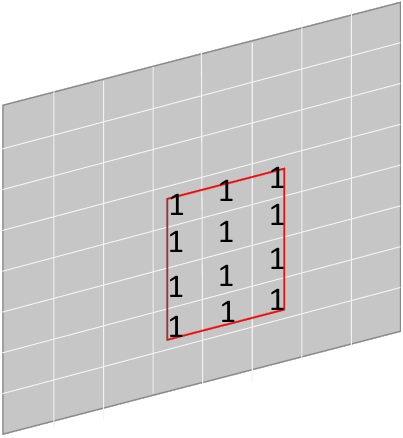}
        \caption{}
        \label{fig:SFTConstruct}
    \end{subfigure}
    \caption{(a) Performance under controlled budget; (b) Illustartions of dense patch-wise label assignment in SFT training.}
    %\label{fig:}
\end{figure}

\noindent \textbf{Ablation studies on threshold $\tau$.} In Sec. 4.4 of the main paper, we present a sensitivity analysis of the selection threshold on the V$^{*}$ benchmark  \cite{wu2024v}. In Fig. \ref{fig:ablateTauAll}, we report the complete experimental results on the V$^{*}$ and HR-Bench 4K/8K benchmarks \cite{wang2025divide}. As shown, TiCO consistently outperforms  Mini-o3 across all threshold values, demonstrating robustness to threshold selection. We adopt $\tau=0.15$, as it achieves consistently strong results across three benchmarks.

\begin{figure*}[t]
	\centering
    \includegraphics[width=\textwidth]{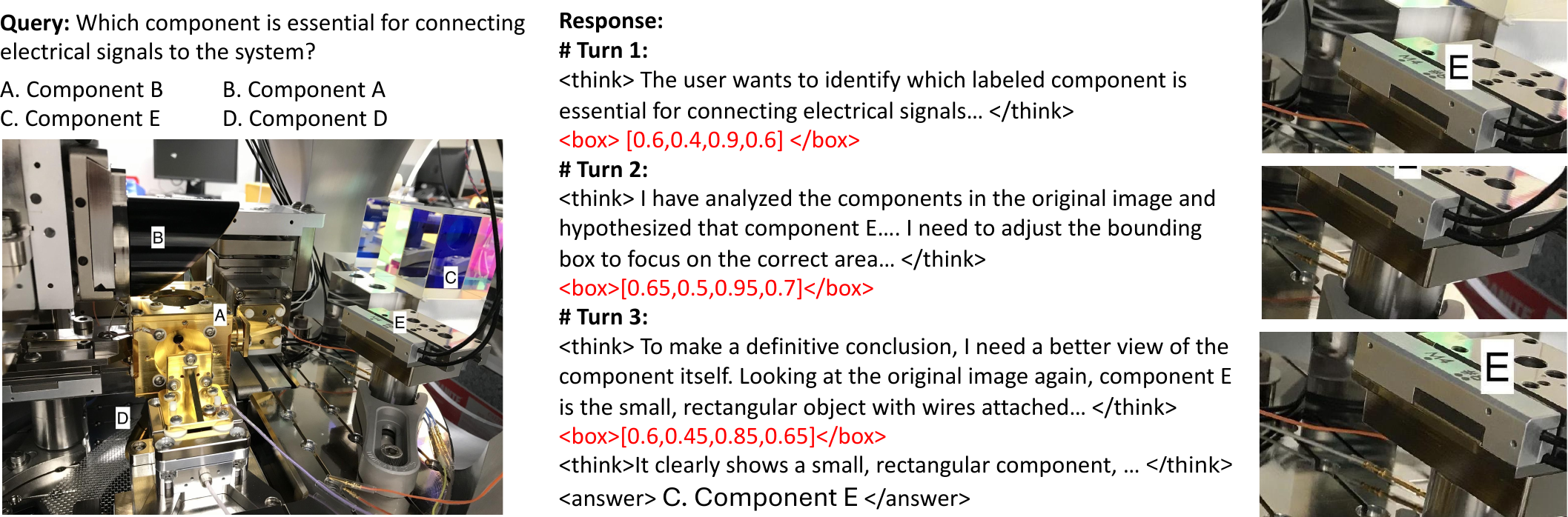}
    % \vspace{-1mm}
	\caption{\textbf{Exemplar SFT dataset} in Mini-o3.}
	\label{fig:SFTDataCase}
	%\vspace{-2mm}
\end{figure*}

\noindent \textbf{Hyper-parameter analysis.} We evaluate the trade-off between the number of evidence regions $N_1$ and answer rollouts $N_2$ under a fixed budget ($N_1 \times N_2 = 8$). As shown in Table \ref{tab:ablationN}, the configuration ($N_1=4, N_2=2$) achieves the best overall performance, suggesting that allocating more capacity to evidence coverage is more beneficial than increasing rollout multiplicity. Allocating more evidence regions improves coverage and grounding reliability, while a moderate number of rollouts remains sufficient for stable answer generation.

% \begin{wrapfigure}[12]{r}{.4\linewidth}
%     \centering
%     \vspace{-8mm}
%     \includegraphics[width=0.8\linewidth]{figs/SFTConstruct.pdf}
%     \vspace{-2mm}
%     \caption{Dense label assignment.}
%     \label{fig:SFTConstruct}
% \end{wrapfigure}

\noindent \textbf{Performance under controlled budget.} To evaluate the reasoning efficiency under controlled per-round token budgets, we constrain the maximum token budget available in each reasoning round to simulate realistic computational limits. Specifically, we vary the per-round token budget for Mini-o3 \cite{lai2025mini} across three settings (256, 128, and 64 tokens) and measure the corresponding accuracy. 

As shown in Figure \ref{fig:budget}, Mini-o3's performance degrades substantially as the per-round token budget decreases. While it achieves 63.0\% accuracy at 256 tokens, the accuracy drops sharply to around 36.1\% at 128 tokens and collapses to nearly 0\% at 64 tokens. This trend indicates a strong reliance on longer reasoning traces and extended generation capacity. In contrast, TiCO attains significantly higher accuracy with a much smaller token allowance, outperforming Mini-o3 even under its most favorable budget setting. These results demonstrate that our TiCO exhibits superior reasoning efficiency and robustness under constrained computational resources, suggesting that its performance gains stem from more compact and effective reasoning rather than simply longer token usage.

\section{Experimental Setups}
\begin{figure*}[t]
	\centering
    \includegraphics[width=0.8\textwidth]{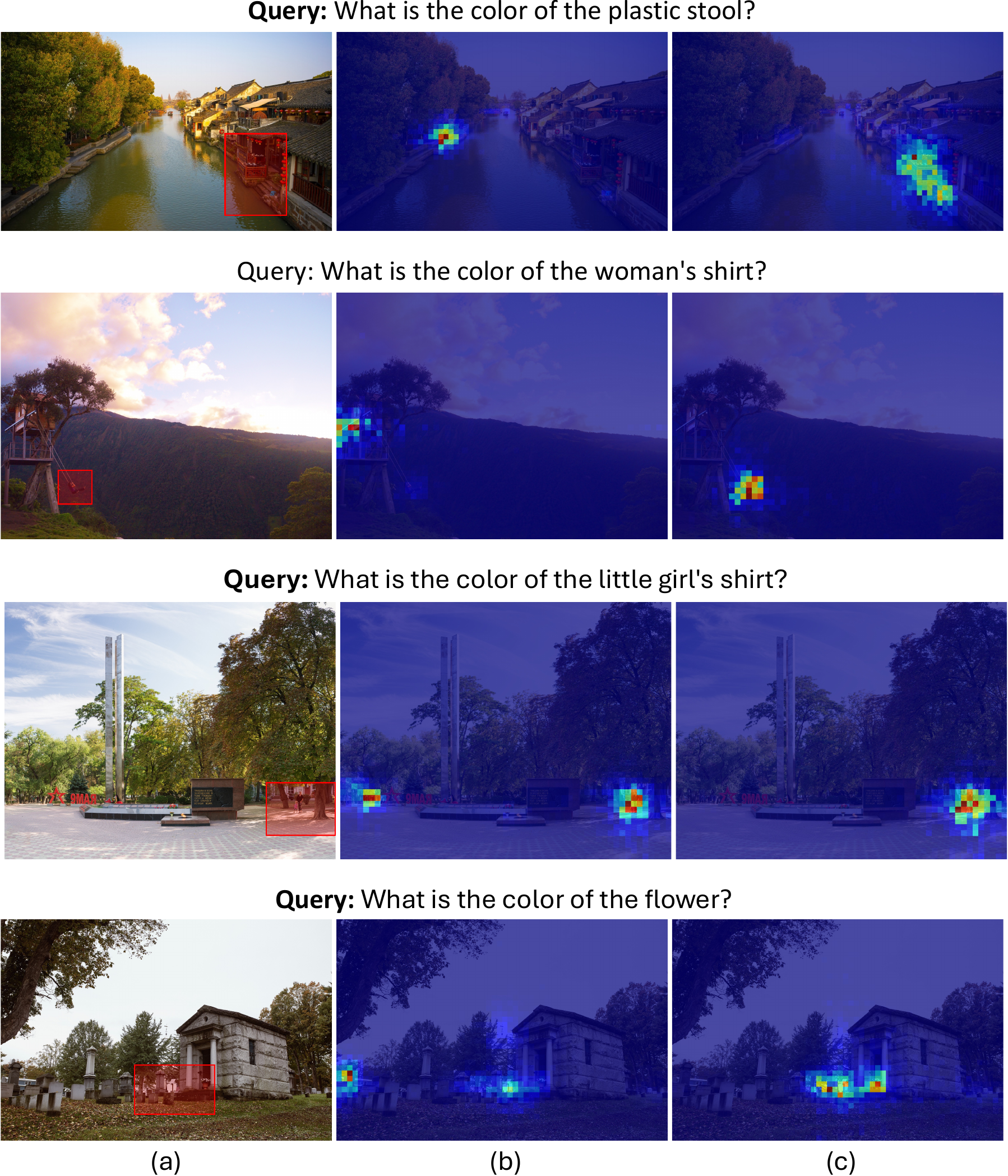}
    \caption{\textbf{Evolution of attention maps from SFT to RL.} (a) Input images with the annotated ground-truth bounding box; (b) Attention map after the SFT stage; (c) Attention map after the RL stage.}
	\label{fig:attentionSFTRL}
	%\vspace{-2mm}
\end{figure*}

\noindent \textbf{Construction details of SFT dataset.} For SFT dataset, we adapt the cold-start SFT dataset introduced in Mini-o3 \cite{lai2025mini} to accommodate our dense similarity-map supervision. As shown in Fig. \ref{fig:SFTDataCase}, the Mini-o3 SFT data contain multi-round refined bounding box predictions, we directly take the final-round bounding box as the ground-truth region. 

Then, we convert the ground-truth bounding box into a binary mask over the image patch grid. Given a normalized bounding box, we scale the normalized coordinates to the patch grid resolution. As shown in Fig \ref{fig:SFTConstruct}, all visual patches that are partially or fully covered by this ground-truth box are labeled as positive.

\noindent \textbf{Evolution of attention maps from SFT to RL.} As illustrated in Fig. \ref{fig:attentionSFTRL}, the visualization of attention maps across training stages reveals a clear evolutionary trajectory from coarse global scanning to fine-grained local focus. In the SFT stage (column (b) in Fig. \ref{fig:attentionSFTRL}), the model's attention is notably diffuse. For instance, in the first row regarding the "\texttt{plastic stool}", the activation is scattered across the river and background architecture, failing to isolate the small-scale target. Conversely, after the RL stage (column (c) in Fig. \ref{fig:attentionSFTRL}), the attention distribution undergoes a significant spatial refinement, exhibiting a sharp convergence that precisely aligns with the ground-truth bounding boxes in column (a). This transition demonstrates that the reinforcement learning process effectively suppresses irrelevant background noise and recalibrates the model's gaze toward the specific discriminative regions.

\noindent \textbf{Robustness Analysis.} To assess the robustness of coordinate-based (\eg, Mini-o3) and coordinate-free methods, we conduct experiments under \emph{noisy} bounding-box supervision (\cf Sec. 4.3 of the main paper). To simulate spatial imprecision, we perturb the canonical bounding box $\mathcal{B} = [x_1, y_1, x_2, y_2]$ by adding zero-mean Gaussian noise scaled by the perturbation factor $\alpha$. Formally, the noisy coordinates are defined as follows.
\begin{equation}
\begin{cases}x_1^{\prime}=x_1+\Delta x_1, & \Delta x_1 \sim \mathcal{N}\left(0,(\alpha \cdot w)^2\right) \\ y_1^{\prime}=y_1+\Delta y_1, & \Delta y_1 \sim \mathcal{N}\left(0,(\alpha \cdot h)^2\right) \\ x_2^{\prime}=x_2+\Delta x_2, & \Delta x_2 \sim \mathcal{N}\left(0,(\alpha \cdot w)^2\right) \\ y_2^{\prime}=y_2+\Delta y_2, & \Delta y_2 \sim \mathcal{N}\left(0,(\alpha \cdot h)^2\right)\end{cases}
\end{equation}
where $\text{side} \in \{w, h\}$ represents the corresponding box dimension. This ensures that the relative shift is proportional to the object size, providing a consistent measure of localization error across varying scales. The corresponding results are shown in Fig. 4 (left) and Sec. 4.3 of the main paper.

%\input{table_figs/figAblateTauAllDataset}

% \noindent \textbf{Ablation studies on threshold $\tau$.} In Sec. 4.4 of the main paper, we present a sensitivity analysis of the selection threshold on the V$^{*}$ benchmark. In Fig. \ref{fig:ablateTauAll}, we report the complete experimental results on the V$^{*}$ and HR-Bench 4K/8K benchmarks. As shown, TiCO consistently outperforms  Mini-o3 across all threshold values, demonstrating robustness to threshold selection. We adopt $\tau=0.15$, as it achieves consistently strong results across three benchmarks.

\begin{figure}[t]
    \centering
    \begin{subfigure}[t]{\linewidth}
        \centering
        \includegraphics[width=\linewidth]{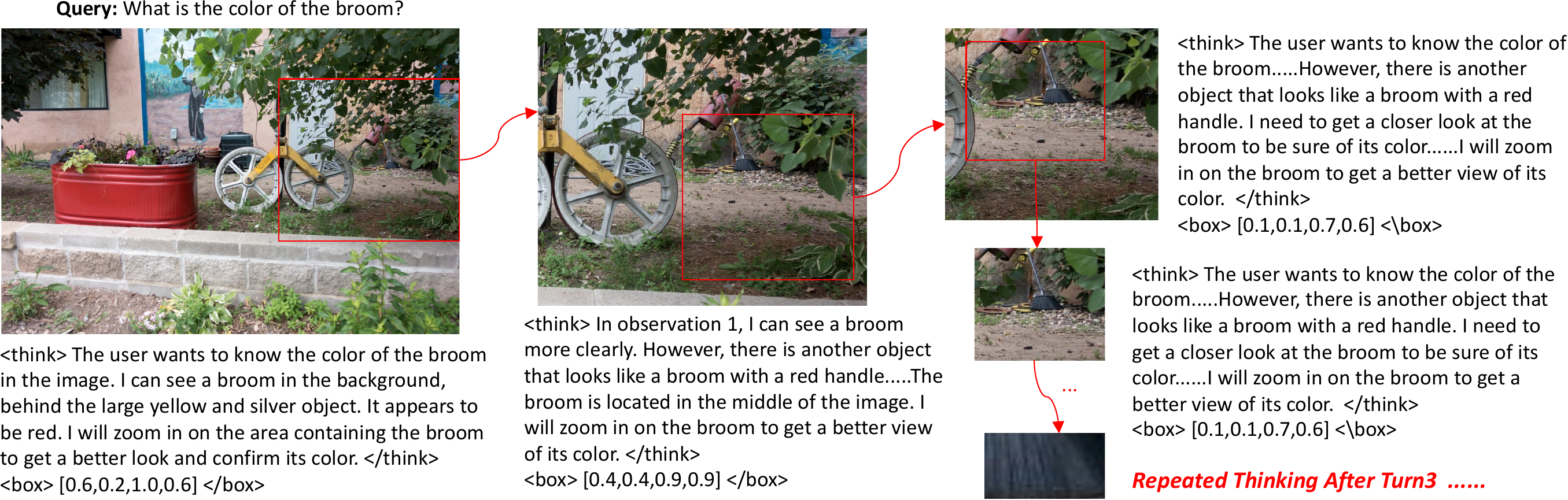}
        \caption{}
        \label{fig:failureModeOfThink1}
    \end{subfigure}
    %\hfill
    \vspace{1mm}
    \begin{subfigure}[t]{\linewidth}
        \centering
        \includegraphics[width=\linewidth]{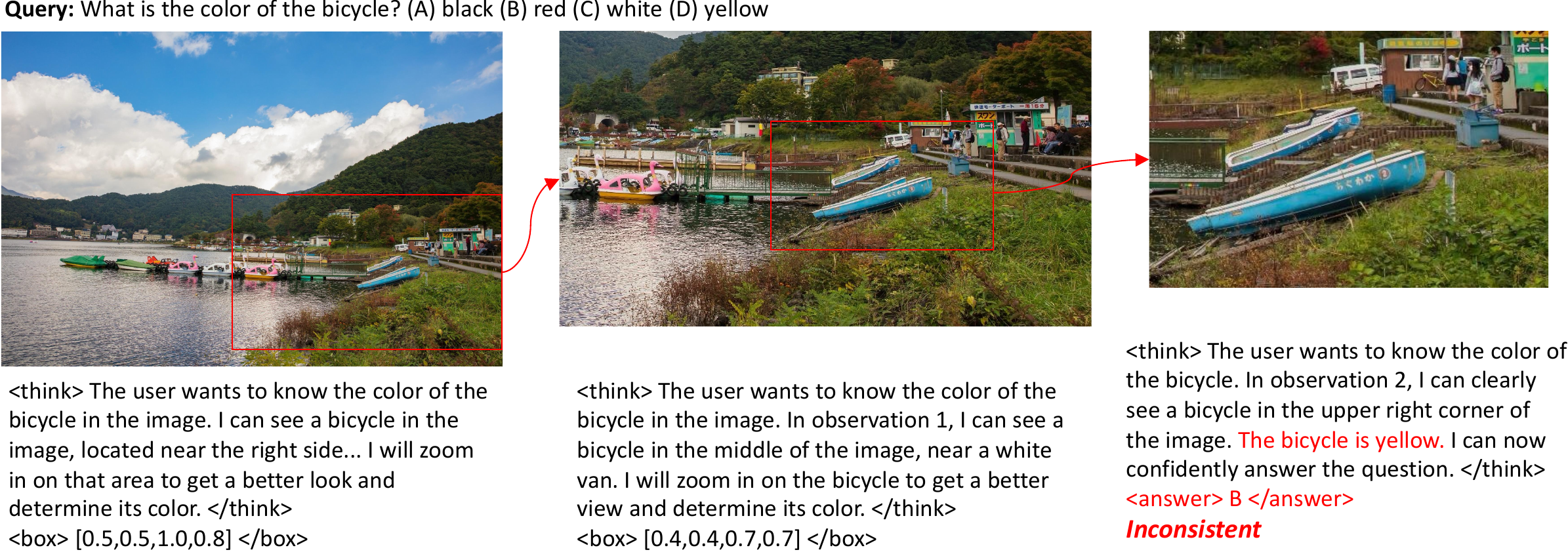}
        \caption{}
        \label{fig:failureModeOfThink2}
    \end{subfigure}
    \caption{\textbf{Qualitative examples of failure cases in the "thinking" mode of coordinate-based methods.} (a) Repeated Thinking: The model falls into an infinite loop during the multi-turn reasoning process, repeatedly generating the same crop coordinates and thoughts without progressing toward a final answer. (b) Inconsistent Logic: Although the model correctly identifies the target object (a yellow bicycle) in the final observation, it produces an incorrect final answer (Option B: red) that contradicts its own internal reasoning and visual evidence.}
    \label{fig:failureModeOfThink}
\end{figure}

\noindent \textbf{Failure examples of the "thinking" mode in coordinate-based methods.} In Table 4 of the main paper, we observe that on certain benchmarks, the coordinate-based variant without thinking slightly outperforms the variant with thinking. We provide qualitative examples in Fig. \ref{fig:failureModeOfThink} to illustrate the common issues that arise in the "thinking" mode of coordinate-based methods.

Typically, we identify two primary bottlenecks: 1) \emph{Recursive Loops:} As shown in Fig. \ref{fig:failureModeOfThink1}, the model often falls into a "dead loop" where it repeatedly generates identical coordinates and justifications (\eg, "need a closer look") without progressing to a conclusion. This suggests a lack of diversity constraints in the iterative sampling process; 2) \emph{Logical Inconsistency:} In Fig. \ref{fig:failureModeOfThink2}, the model achieves correct perception (stating "the bicycle is yellow") through precise zooming, yet contradicts itself by selecting an incorrect final answer (Option B: red). This indicates a failure to propagate correctly perceived high-resolution features into the final decision head.

\begin{table*}[t]
\centering
\caption{\textbf{Prompts for TiCO.} The placeholder \textcolor{red}{\{input\}} is dynamically instantiated with the task-specific query and the corresponding high-resolution image context during both the training and inference phases.} 
\vspace{-2mm}
    %\begin{tabu}[h]{0.95\textwidth}
        \centering
        %\resizebox{\textwidth}{!}{
        \begin{tabular}{p{13cm}}
        \hline
        Question: \textcolor{red}{\{input\}} \\
        Your task is to locate the region that corresponds to the question. To indicate the region, we will use some special tokens, which is used to refer to a visual patch later. For example, you can output: $<$your\_special\_token\_here$>$.\\
        Refer to the region extracted in the previous turn and provide the final answer.\\
        %Then, you will receive a cropped image based on the bounding box. Use all the images to continue reasoning. You may conduct multiple rounds of grounding to refine your region as you want. The bounding box you provide should always be selected based on the original image. If at any point you determine no further visual information is needed, you may directly provide the final answer inside the \answer{and} tags.\\
        %Format Example:\\
        %\think{Reasoning Process} \bbox{[x1,y1,x2,y2]} or, \\
        %\think{Reasoning Process}\answer{Final Answer} \\ %\answer{Final Answer} \\
        \hline
        \end{tabular}
        %}
        %\caption{}
        %\label{tab:groundPrompt}
     %\end{subtable} %\vspace{15mm} %\hfill
     \label{tab:Prompt}
%\vspace{-1em}
\end{table*}

\noindent \textbf{Prompt.} As detailed in Table \ref{tab:Prompt}, we employ a structured prompt template to guide the multi-turn reasoning process of TiCO. The placeholder \textcolor{red}{\{input\}} is dynamically instantiated with the task-specific query and the corresponding high-resolution image context during both the training and inference phases. Following the initial query, the prompt explicitly instructs the MLLM to perform a two-step reasoning task: first, to locate the relevant visual region using special latent tokens and second, to synthesize the final answer by referring back to the extracted evidence from previous turns. 

\noindent \textbf{Details of connected component generation.} Given the similarity (or attention) score $\alpha_i$ for each visual token $v_i$, we first reshape the token sequence back to its original 2D spatial layout according to the patch grid of the vision encoder (\eg, a $H \times W$ grid for ViT-style encoders). We then apply a threshold $\tau$ to obtain a binary mask $\mathcal{M}$.

Connected components are defined over the 2D token grid based on spatial adjacency inherited from the vision encoder. We adopt \emph{4-neighborhood connectivity}, where two tokens are considered connected if they share a horizontal or vertical boundary (\ie, top, bottom, left, or right neighbors). Diagonal connections are not included. Formally, for tokens located at grid positions $(x,y)$ and $(x^{\prime}, y^{\prime})$, they are adjacent if:
\begin{equation}
\left|x-x^{\prime}\right|+\left|y-y^{\prime}\right|=1.
\end{equation}
For each connected component $\mathcal{C}_k$, we compute its minimal enclosing rectangle in the original image coordinate system. Specifically, let $(x_{min},y_{min})$ and $(x_{max},y_{max})$ denote the extreme grid coordinates of tokens in $\mathcal{C}_k$. The corresponding image-space bounding rectangle is obtained by mapping patch indices back to pixel coordinates using the patch size and input resolution. 

Importantly, these rectangles are not predicted by the model and do not serve as regression targets. They are deterministically derived from the similarity-induced token clusters and are used solely as an evidence extraction interface.
\begin{figure*}[t]
	\centering
    \includegraphics[width=0.9\textwidth]{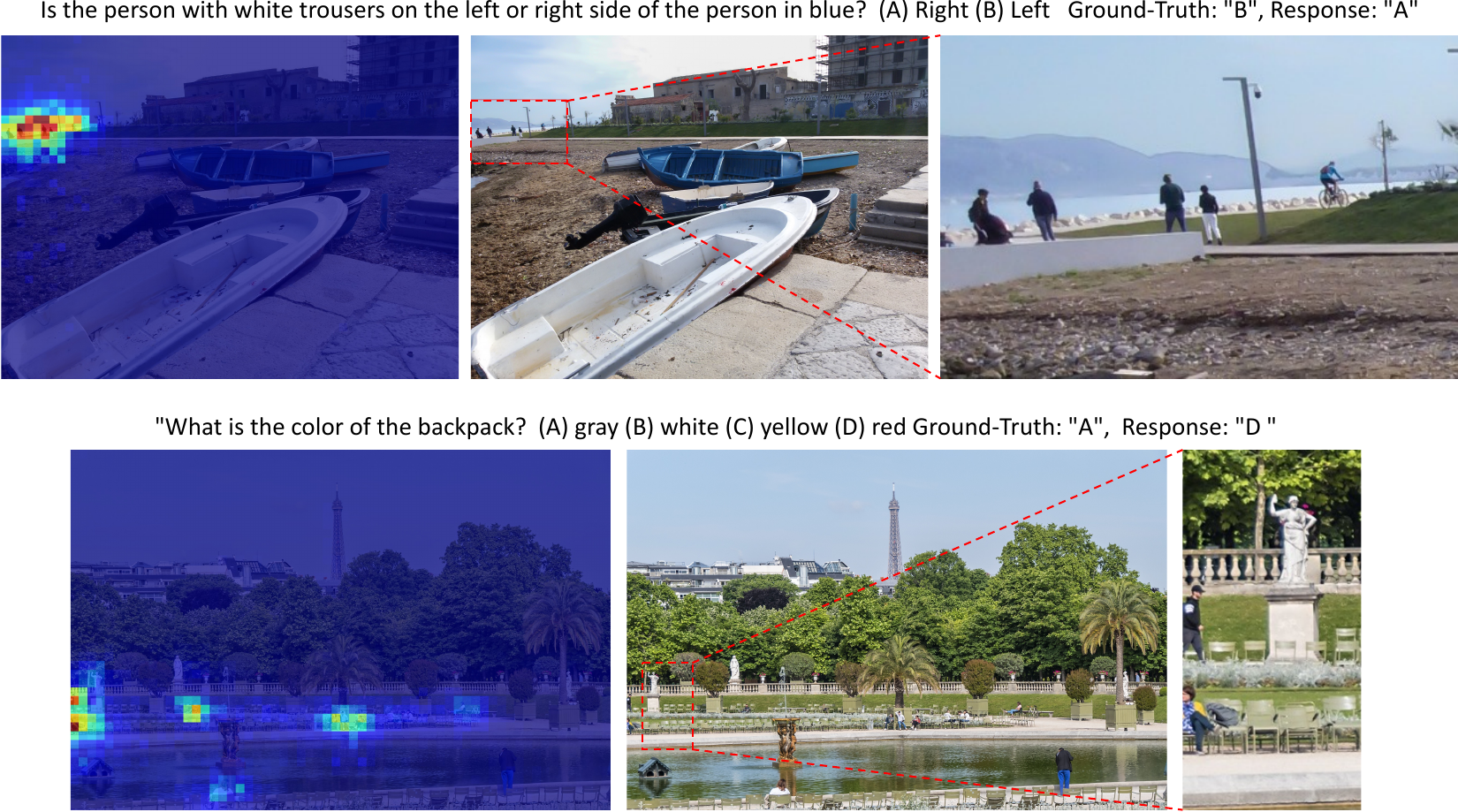}
    % \vspace{-1mm}
	\caption{\textbf{Failure cases of our TiCO.} (Top) Spatial relationship error: TiCO incorrectly predicts the left–right relation between two persons, revealing limitations in spatial reasoning. (Bottom) Detailed attribution error: TiCO misidentifies the backpack color (predicting red instead of gray), indicating weaknesses in fine-grained attribute recognition.}
	\label{fig:FailureCase}
	%\vspace{-2mm}
\end{figure*}

\noindent \textbf{Limitations.} Despite its empirical effectiveness and efficiency advantages, our TiCO has three limitations. First, although TiCO eliminates explicit coordinate regression as a modeling objective, it still relies on threshold-based binarization, which may require the manual selection of the threshold. Second, TiCO does not employ an explicit \emph{multi-step} "thinking" process during reasoning. While this design substantially improves efficiency by avoiding long-form auto-regressive reasoning, it also removes the reflective characteristics often observed in thinking-based reasoning frameworks, such as the iterative refinement or "Aha moment" behavior reported in R1-style models. As a result, TiCO favors streamlined evidence-driven decision-making over self-corrective reasoning dynamics, which may limit its adaptability in tasks requiring complex deliberation or multi-hop reflection. Finally, as a general challenge for thinking-with-images paradigms, high-resolution reasoning remains sensitive to grounding quality. If the selected sub-region fails to accurately cover the semantically relevant evidence, subsequent reasoning, regardless of whether it is coordinate-based or coordinate-free, may struggle to recover the correct answer.
\begin{figure}[t]
    \centering
    \begin{subfigure}[t]{0.96\linewidth}
        \centering
        \includegraphics[width=0.9\linewidth]{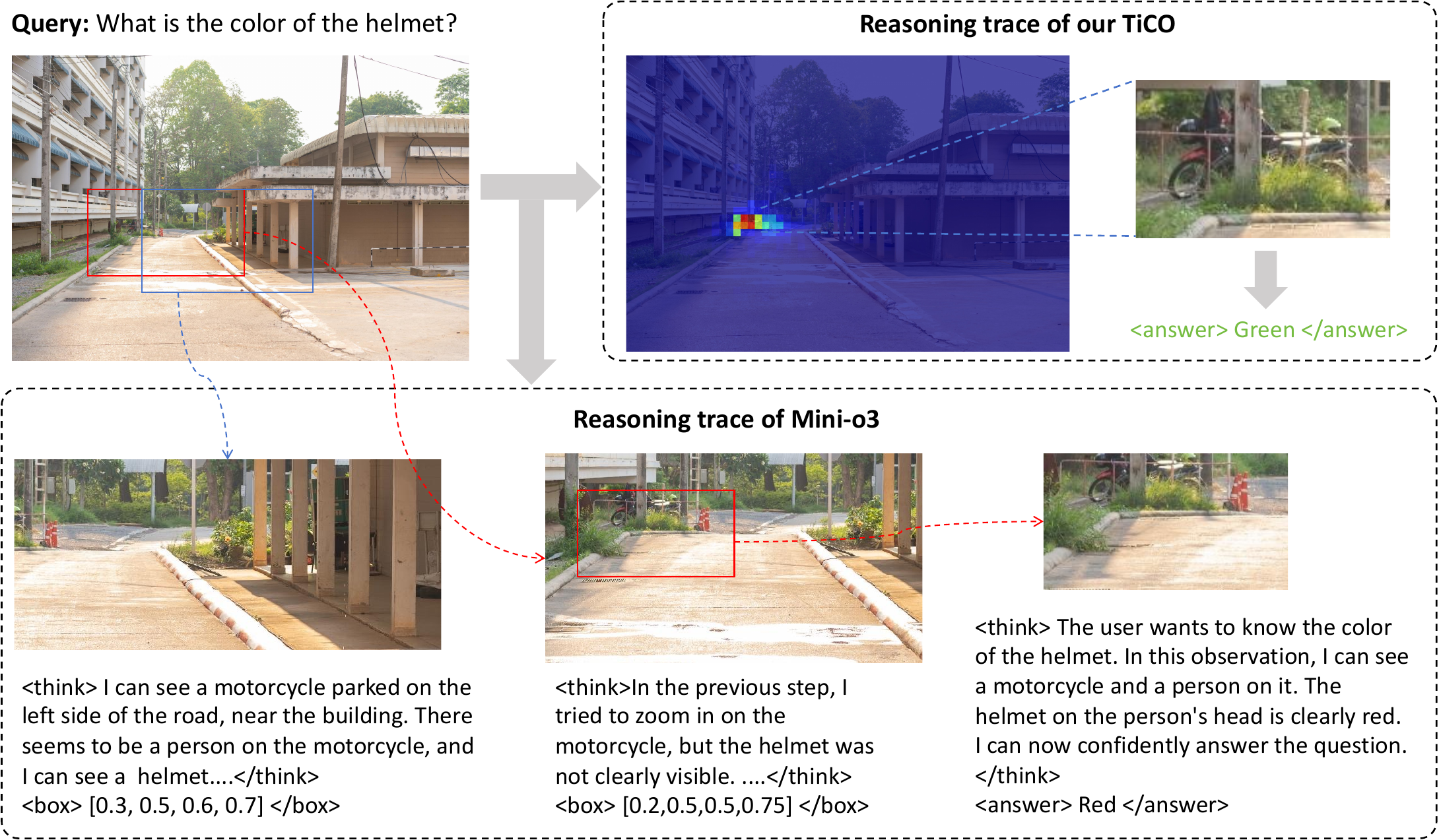}
        \caption{}
        \label{fig:visCompare1}
    \end{subfigure}
    %\hfill
    \vspace{1mm}
    \begin{subfigure}[t]{0.96\linewidth}
        \centering
        \includegraphics[width=0.9\linewidth]{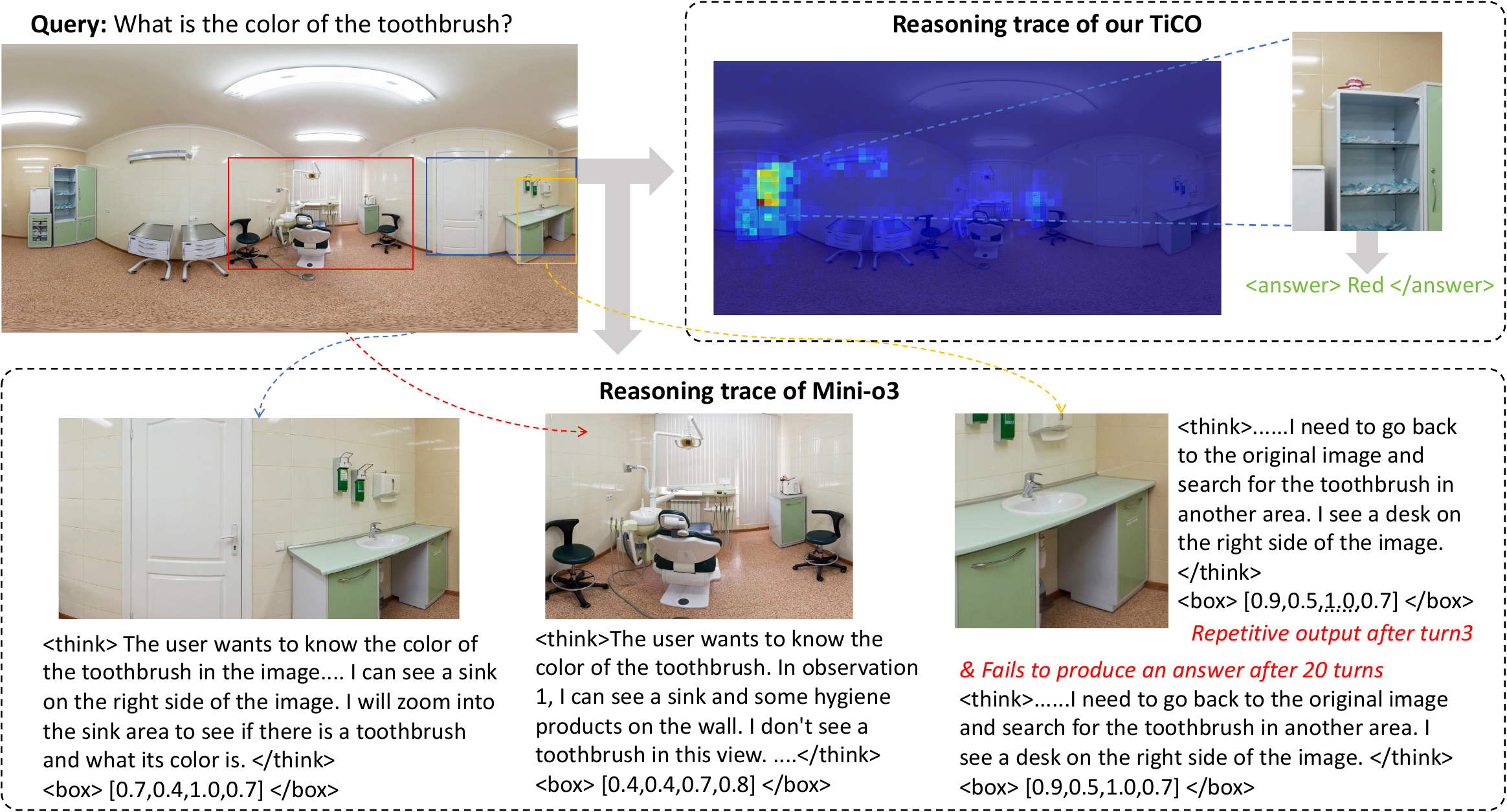}
        \caption{}
        \label{fig:visCompare2}
    \end{subfigure}
    \caption{Comparison between coordinate-based methods and TiCO.}
    \label{fig:visCompare}
\end{figure}

\section{Visualization}

\noindent \textbf{Failure cases of our TiCO.} To further investigate the limitations of the proposed TiCO model, we conduct a qualitative analysis of its failure cases, specifically focusing on spatial reasoning and fine-grained attribute recognition. As illustrated in Fig. \ref{fig:FailureCase}, two primary types of errors emerge: 1) Spatial relationship misjudgment: In the top row of Fig. \ref{fig:FailureCase}, the model is tasked with determining the left-right spatial relationship between two individuals. Although TiCO successfully utilizes its grounding mechanism to zoom in on the relevant subjects (as evidenced by the high-resolution crop), it fails to correctly interpret their relative positioning, providing an answer of "Right" instead of the ground-truth "Left". This suggests that even when the model achieves precise visual alignment with the target objects, its internal reasoning module may still struggle with complex geometric or relational logic in coordinate-free spaces; 2) Fine-grained attribute recognition errors: The bottom row of Fig. \ref{fig:FailureCase} demonstrates a failure in detailed attribution, specifically regarding the color of a small object (a backpack). Despite the model's attention heatmaps indicating successful localization of the target area, the model incorrectly predicts the color as "red" instead of "gray". This indicates that while TiCO effectively narrows the "resolution gap" through targeted zooming, it occasionally fails to extract or propagate the correct semantic features needed for high-fidelity attribute classification.

\noindent \textbf{Comparison between coordinate-based methods and TiCO.} Fig. \ref{fig:visCompare} provides qualitative comparisons between coordinate-based methods Mini-o3 \cite{lai2025mini} and our coordinate-free TiCO under high-resolution reasoning scenarios. In the first example, TiCO directly concentrates its attention on the semantically relevant region containing the motorcycle helmet and correctly identifies the answer ("Green"), while Mini-o3 relies on explicit coordinate regression and produces misaligned crops, ultimately leading to an incorrect prediction ("Red"). This illustrates that coordinate-based methods are sensitive to localization errors, where inaccurate bounding boxes can propagate through subsequent reasoning steps. In the second example, the discrepancy becomes more pronounced. TiCO places its semantic attention over connected regions and successfully identifies the toothbrush, producing the correct answer. In contrast, Mini-o3 repeatedly generates bounding boxes over irrelevant regions and enters a repetitive reasoning loop without producing a valid answer after multiple turns. This behavior highlights a structural weakness of coordinate-based grounding, where bounding box prediction may lead to unstable exploration and error accumulation.

% \noindent \textbf{Mini-o3 performance under controlled budget.}

% \noindent \textbf{Open-source.}

% \textcolor{red}{--------}

% Anonymous Github

% What if the selected regions incorrect?

% Mini-o3 performance under controlled budget

% attention map (after SFT/RL) \checkmark

% Failure Case (multiple objects, spatial relationships) \checkmark

% Prompts \checkmark

% thinking vs non-thinking (1. Inconsistent; 2. Replication) \checkmark

% LIMITATIONS \checkmark

% Supplementary: our SFT construction details \checkmark

% Robust Analysis (Define alpha) \checkmark

% connected-component generation manner \checkmark

% Ablation studies on threshold \checkmark

% Comparative visualizations \checkmark

% \textcolor{red}{--------}

%\clearpage  % TODO FINAL: This \clearpage needs to be removed from both review and camera-ready versions.

% \section*{Acknowledgements}
% Please insert your acknowledgments here.


\end{document}